\documentclass{article}

\usepackage{microtype}
\usepackage{graphicx}
\usepackage{subfigure}
\usepackage{booktabs} 

\usepackage{hyperref}

\usepackage[accepted]{icml2020}

\usepackage{amsmath}
\usepackage[export]{adjustbox}
\usepackage{tikz}
\usetikzlibrary{positioning,calc,bayesnet,external}
\renewcommand{\algorithmicrequire}{\textbf{Input:}}
\renewcommand{\algorithmicensure}{\textbf{Output:}}

\newcommand{\ourLCP}{LCRankNet}

\icmltitlerunning{Learning to Rank Learning Curves}

\begin{document}

\twocolumn[
\icmltitle{Learning to Rank Learning Curves}

\icmlsetsymbol{equal}{*}

\begin{icmlauthorlist}
\icmlauthor{Martin Wistuba}{aff1}
\icmlauthor{Tejaswini Pedapati}{aff1}
\end{icmlauthorlist}

\icmlaffiliation{aff1}{IBM Research}

\icmlcorrespondingauthor{Martin Wistuba}{martin.wistuba@ibm.com}

\icmlkeywords{Automated Machine Learning, Neural Architecture Search, Learning Curve Prediction}

\vskip 0.3in
]

\printAffiliationsAndNotice{}  

\begin{abstract}
Many automated machine learning methods, such as those for hyperparameter and neural architecture optimization, are computationally expensive because they involve training many different model configurations.
In this work, we present a new method that saves computational budget by terminating poor configurations early on in the training.
In contrast to existing methods, we consider this task as a ranking and transfer learning problem.
We qualitatively show that by optimizing a pairwise ranking loss and leveraging learning curves from other datasets, our model is able to effectively rank learning curves without having to observe many or very long learning curves.
We further demonstrate that our method can be used to accelerate a neural architecture search by a factor of up to 100 without a significant performance degradation of the discovered architecture.
In further experiments we analyze the quality of ranking, the influence of different model components as well as the predictive behavior of the model.
\end{abstract}

\section{Introduction}
A method commonly used by human experts to speed up the optimization of neural architectures or hyperparameters is the early termination of iterative training processes that are unlikely to improve the current solution.
A common technique to determine the likelihood of no improvement is to compare the learning curve of a new configuration to the one of the currently best configuration.
This idea can also be used to speed up automated machine learning processes.
For this purpose, it is common practice to extrapolate the partial learning curve in order to predict the final performance of the currently investigated model.
Current extrapolation techniques have several weaknesses that make them unable to realize their full potential in practice.
Many of the methods require sufficient sample learning curves to make reliable predictions~\citep{Chandrashekaran2017,Klein2017,Baker2018}.
Thus, the extrapolation method for the first candidates can not be used yet, which means more computational effort.
Other methods do not have this disadvantage, but require sufficiently long learning curves to make reliable predictions which again means unnecessary overhead~\citep{Domhan2015}.
Many of these methods also do not take into account other information such as the hyperparameters of the model being examined or its network architecture.

We address the need for sample learning curves by devising a transfer learning technique that uses learning curves from other problems.
Since the range of accuracy varies from dataset to dataset, we are forced to consider this in our modeling. 
But since we are not interested in predicting the performance of a model anyway, we use a ranking model that models the probability that the model currently being investigated surpasses the best solution so far.

In order to be able to make reliable predictions for short learning curves, we consider further characteristics of the model such as its network architecture. 
We compare our ranking method with respect to a ranking measure against different methods on five different image classification and four tabular regression datasets.
We also show that our method is capable of significantly accelerating neural architecture search (NAS) and hyperparameter optimization.
Furthermore, we conduct several ablation studies to provide a better motivation of our model and its behavior.

\section{Related work}
Most of the prior work for learning curve prediction is based on the idea of extrapolating the partial learning curve by using a combination of continuously increasing basic functions.

\citet{Domhan2015} define a set of 11 parametric basic functions, estimate their parameters and combine them in an ensemble.
\citet{Klein2017} propose a heteroscedastic Bayesian model which learns a weighted average of the basic functions.
\citet{Chandrashekaran2017} do not use basic functions but use previously observed learning curves of the current dataset.
An affine transformation for each previously seen learning curve is estimated by minimizing the mean squared error with respect to the partial learning curve.
The best fitting extrapolations are averaged as the final prediction.
\citet{Baker2018} use a different procedure.
They use support vector machines as sequential regressive models to predict the final accuracy based on features extracted from the learning curves, its gradients, and the neural architecture itself.

The predictor by \citet{Domhan2015} is able to forecast without seeing any learning curve before but requires observing more epochs for accurate predictions.
The model by \citet{Chandrashekaran2017} requires seeing few learning curves to extrapolate future learning curves.
However, accurate forecasts are already possible after few epochs.
Algorithms proposed by \citet{Klein2017,Baker2018} need to observe many full-length learning curves before providing any useful forecasts.
However, this is prohibiting in the scenarios where learning is time-consuming such as in large convolutional neural networks (CNN).

All previous methods for automatically terminating iterative learning processes are based on methods that predict the learning curve.
Ultimately, however, we are less interested in the exact learning curve but rather whether the current learning curve leads to a better result.
This way of obtaining a ranking is referred to as pointwise ranking methods~\citep{Liu2011}.
They have proven to be less efficient than pairwise ranking methods which directly optimize for the objective function~\citep{Burges2005}.
Yet, we are the first to consider a pairwise ranking loss for this application.

There are some bandit-based methods which leverage early termination as well.
Successive Halving~\citep{Jamieson2016} is a method that trains multiple models with settings chosen at random simultaneously and terminates the worst performing half in predefined intervals.
Hyperband~\citep{Li2017} identifies that the choice of the intervals is vital and therefore proposes to run Successive Halving with different intervals.
BOHB~\citep{Falkner2018} is an extension of Hyperband which proposes to replace the random selection of settings with Bayesian optimization.

The use of methods that terminate less promising iterative training processes early are of particular interest in the field of automated machine learning, as they can, for example, significantly accelerate hyperparameter optimization.
The most computationally intensive subproblem of automated machine learning is Neural Architecture Search~\citep{Wistuba2019}, the optimization of the neural network topology.
Our method is not limited to this problem, but as it is currently one of the major challenges in automated machine learning, we use this problem as a sample application in our evaluation.
For this particular application, optimization methods that leverage parameter sharing~\citep{Pham2018} have become established as a standard method.
Here, a search space spanning, overparameterized neural network is learned and finally used to determine the best architecture.
DARTS~\citep{Liu2019} is one of the last extensions of this idea which uses a continuous relaxation of the search space, which allows learning of the shared parameters as well as of the architecture by means of gradient descent.
These methods can be a strong alternative to early termination but do not transfer to general machine learning and recent research implies that this approach does not work for arbitrary architecture search spaces~\citep{Yu2020}.

\section{Learning curve ranking}
\begin{figure}
  \begin{center}
    \resizebox{1\columnwidth}{!}{\begin{tikzpicture}
  \draw[->] (0,0) -- (10,0) node[below,pos=0.5]{\large Iterations};
  \draw[->] (0,0) -- (0,4.2) node[above,pos=0.5,rotate=90]{\large Validation Accuracy};
  \draw[scale=0.5,domain=0:2,smooth,variable=\x,black] plot ({\x},{10-6/ln(\x+2)});
  \draw[scale=0.5,domain=2:19,dashed,variable=\x,black] plot ({\x},{10-6/ln(\x+2)});
  \node at (9.5,4) (mp-cross) {$\times$} ;
  \node[below=0.01 of mp-cross] {\large Final Performance};
  \node at (1.1,2.9) {$\times$} ;
  \node at (3.2,2.6) {\large Intermediate Performance};
\end{tikzpicture}}
  \end{center}
  \caption{Learning curve prediction tries to predict from the partial learning curve (solid line) the final performance.}
  \label{fig:early-termination-learning-curve}
\end{figure}
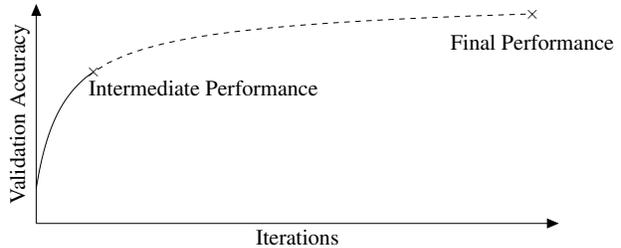
With \emph{learning curve} we refer to the function of qualitative performance with growing number of iterations of an iterative learning algorithm.
We use the term \emph{final learning curve} to explicitly denote the entire learning curve, $y_1,\ldots,y_L$, reflecting the training process from beginning to end.
Here, $y_i$ is a measure of the performance of the model (e.g., classification accuracy), which is determined at regular intervals.
On the contrary, a \emph{partial learning curve}, $y_1,\ldots,y_l$, refers to learning curves that are observed only up to a time $l$.
We visualize the concepts of the terms in Figure~\ref{fig:early-termination-learning-curve}.

There is a set of automated early termination methods which follow broadly the same principle.
The first model is trained to completion and is considered as the current best model $m^{\text{max}}$ with its performance being $y^{\text{max}}$.
For each further model $m_i$, the probability that it is better than the current best model, $p\left(m_i>m^{\text{max}}\right)$,  is monitored at periodic intervals during training.
If it is below a given threshold, the model's training is terminated (see Algorithm~\ref{alg:early-stopping}).
All existing methods rely on a two-step approach to determine this probability which involves extrapolating the partial learning curve and a heuristic measure.
Instead of the two-step process to determine the probability, we propose \ourLCP{} to predict the probability that a model $m_i$ is better than $m_j$ directly.
\ourLCP{} is based on a neural network $f$ which considers model characteristics and the partial learning curve $\mathbf{x}_i$ as input.
We define the probability that $m_{i}$ is better than $m_{j}$ as
\begin{equation}
p(m_i>m_j)=\hat{p}_{i,j}=\frac{e^{f(\mathbf{x}_i)-f(\mathbf{x}_j)}}{1+e^{f(\mathbf{x}_i)-f(\mathbf{x}_j)}}.
\label{eq:model-prediction}
\end{equation}
Using the logistic function is an established modelling approach in the learning-to-rank community~\citep{Burges2005}.
Given a set of final learning curves for some models, the estimation of the values $p_{i,j}$ between these models is trivial since we know whether $m_i$ is better than $m_j$ or not.
Therefore, we set
\begin{equation}
p_{i,j}=\begin{cases}
1 & \text{if }m_{i}>m_{j}\\
0.5 & \text{if }m_{i}=m_{j}\\
0 & \text{if }m_{i}<m_{j}\ .
\end{cases}
\end{equation}
We minimize the cross-entropy loss
\begin{equation}
\label{eq:pairwise-ranking-loss}
L_{\text{ce}} = \sum_{i,j}-p_{i,j}\log\hat{p}_{i,j}-(1-p_{i,j})\log(1-\hat{p}_{i,j})
\end{equation}
to determine the parameters of $f$.
Now $p(m_i>m_j)$ can be predicted for arbitrary pairs $m_i$ and $m_j$ using Equation \eqref{eq:model-prediction}.
A model $f_l$ is trained on all partial learning curves of length $l$ to predict the probabilities of curves of length $l$.
\begin{algorithm}
\caption{Early Termination Method}
\label{alg:early-stopping}
\begin{algorithmic}[1]
\REQUIRE{Dataset $d$, model $m$, performance of best model $m^{\text{max}}$ so far $y^{\text{max}}$}.
\ENSURE{Learning curve.}
\FOR{$l\leftarrow 1\ldots L$}
\STATE Train $m$ on $d$ for a step and observe a further part of the learning curve $y_{t}$.
\IF{$\max_{1\leq i\leq l}y_{i}>y^{\text{max}}$}
  \STATE \textbf{continue}
\ELSIF{$p\left(m>m^{\text{max}}\right) \leq \delta$}
  \STATE \textbf{return} $\mathbf{y}$
\ENDIF
\ENDFOR
\STATE \textbf{return} $\mathbf{y}$
\end{algorithmic}
\end{algorithm}

\begin{figure}
    \begin{center}
        \resizebox{0.8\columnwidth}{!}{\begin{tikzpicture}
\definecolor{colorSegment}{HTML}{ffec9e}
\definecolor{colorCell}{HTML}{ddff88}
\definecolor{colorReductionCell}{HTML}{ffdd88}
\definecolor{colorOp}{HTML}{a6f4fd}
\definecolor{colorBlock}{HTML}{bbee66}

\tikzstyle{boxstyle}=[rectangle,node distance=1.2cm,rounded corners=1ex,align=center]
\tikzstyle{opstyle}=[boxstyle,draw=black,fill=colorOp, minimum size=7mm,align=center]
\tikzstyle{concatstyle}=[opstyle]
\tikzstyle{segmentstyle}=[boxstyle,draw=black,fill=colorSegment]
\tikzstyle{cellstyle}=[boxstyle,draw=black,fill=colorCell,align=center]

  \tikzstyle{oprotatedstyle}=[opstyle,node distance=1.3cm]
  \tikzstyle{cellrotatedstyle}=[cellstyle,node distance=1.7cm]
  
  \tikzstyle{cellstyle-t}=[segmentstyle,node distance=1cm,fill=colorCell]
  \tikzstyle{opstyle-t}=[opstyle,node distance=1cm]
  
  \node[boxstyle](output) at (0,5) {$f(\mathbf{x})$};

  \node[oprotatedstyle, minimum width=7.5cm,align=center,text width=5cm](Dense)[below of= output] {Fully Connected Layer};
  
  \draw[->](Dense) -- (output);
  
   \node[oprotatedstyle](maxpool-2) [below left=0.5cm and -3cm of Dense] {max\\ pool};
   
  \node[oprotatedstyle](conv-2) [below of=maxpool-2] {conv\\k=2..5};
  
  \draw[->]   (conv-2) -- (maxpool-2);
  
  \draw[->] (maxpool-2.north)--(maxpool-2 |- Dense.south);
  
 \node[oprotatedstyle,align=center,fill=gray!10](lcp-input)[below of=conv-2] {$y_{1,\ldots,l}$};

   \node[oprotatedstyle](lstm) [left=0.3cm and 0.7cm of maxpool-2] {LSTM};
 
  \node[oprotatedstyle](embedding) [below of=lstm] {Embedding};
  
   \node[oprotatedstyle,fill=gray!10](arch)[below of=embedding] {Architecture};
   
  \draw[->]   (arch) -- (embedding);
  \draw[->]   (embedding) -- (lstm);
 
 \draw[->] 
  (lstm.north)--(lstm |- Dense.south);

\node[oprotatedstyle](embed-layer) [right=0.3cm and 0.3cm of conv-2] {Embedding};

   \node[oprotatedstyle,fill=gray!10](taskid)[below of=embed-layer] {Dataset ID};
   \draw[->]   (taskid) -- (embed-layer);
   \draw[->] 
  (embed-layer.north)--(embed-layer |- Dense.south);
   
\node[oprotatedstyle,fill=gray!10](hyperparams)[right=0.7cm and 0.3cm of taskid] {Hyperparams};
\draw[->] (hyperparams.north)--(hyperparams |- Dense.south);

  \draw[->]  (lcp-input) -- (conv-2);

\end{tikzpicture}}
    \end{center}
    \caption{\ourLCP{} has four different components, each dealing with one type of input: architecture encoding, partial learning curve, dataset ID, and further hyperparameters.}
    \label{fig:ourLCP-architecture}
\end{figure}
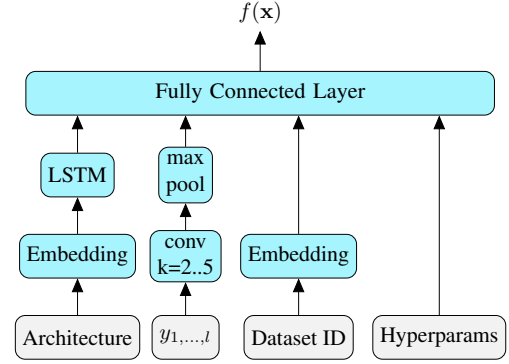
\subsection{Ranking model to learn across datasets}
We have defined the prediction of our model in Equation~\eqref{eq:model-prediction}.
It depends on the outcome of the function $f$ which takes a model representation $\mathbf{x}$ as input.
The information contained in this representation depends on the task at hand.
The representation consists of up to four different parts.
First, the partial learning curve $y_1,\ldots,y_l$.
Second, the description of the model's architecture which is a sequence of strings representing the layers it is comprised of.
Third, a dataset ID to indicate on which dataset the corresponding architecture and learning curve was trained and observed.
Fourth, all remaining hyperparameters, e.g. learning rate or batch size.

We model $f$ using a neural network and use special layers to process the different parts of the representation.
The learned representation for the different parts are finally concatenated and fed to a fully connected layer.
The architecture is visualized in Figure~\ref{fig:ourLCP-architecture}.
We will now describe the different components in detail.

\begin{figure*}[t]
\centering
\includegraphics[width=0.32\textwidth]{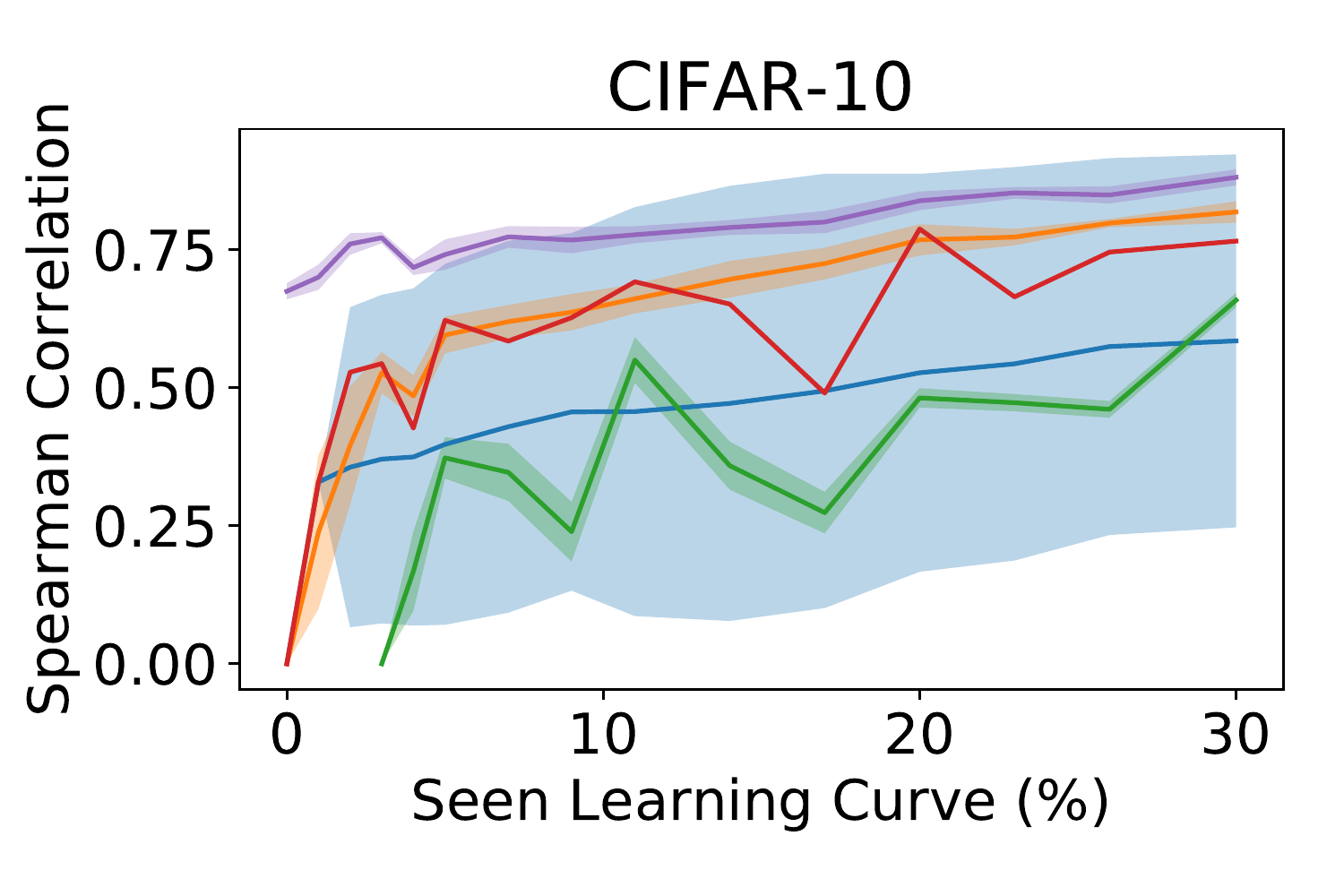}
\includegraphics[width=0.32\textwidth]{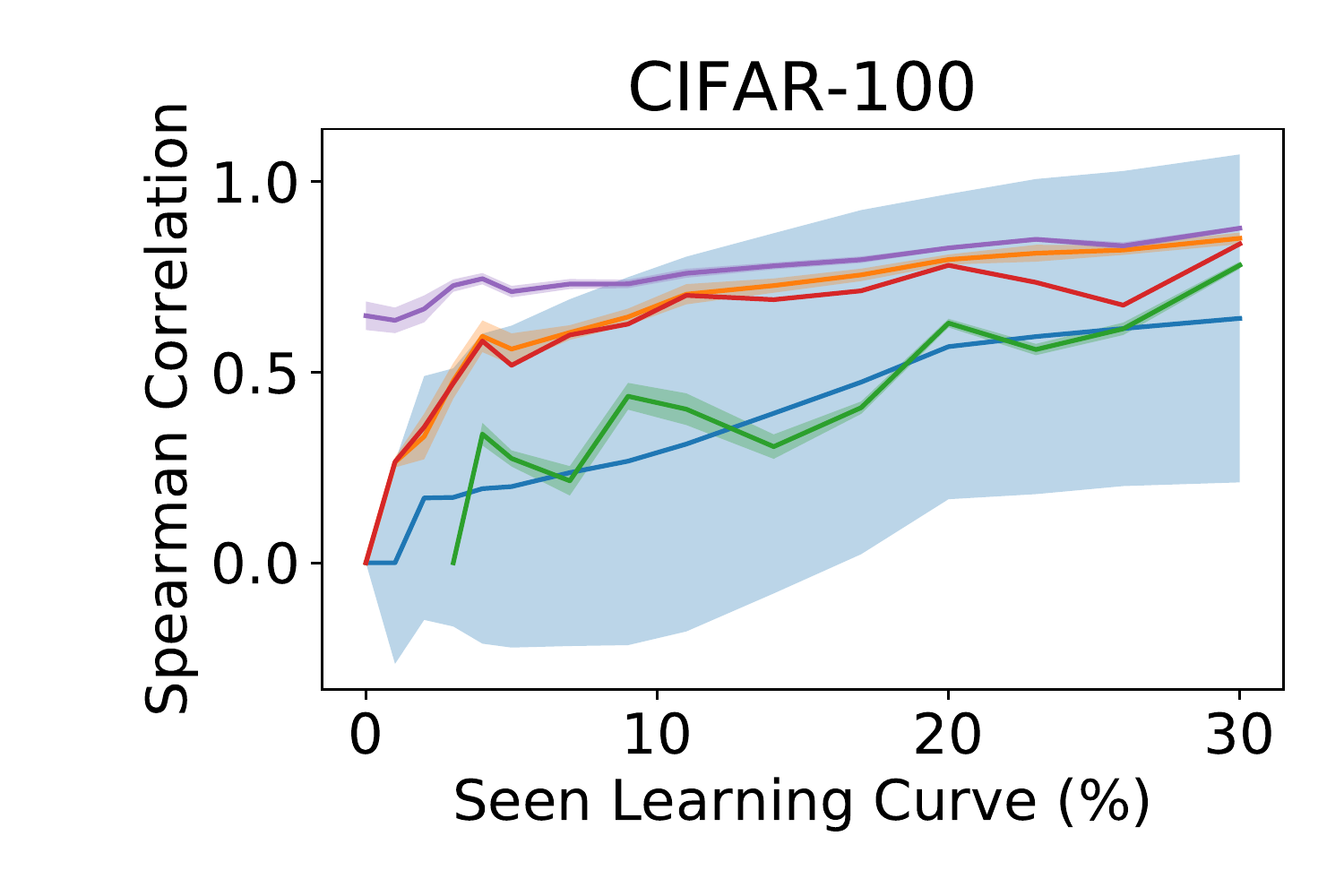}
\includegraphics[width=0.32\textwidth]{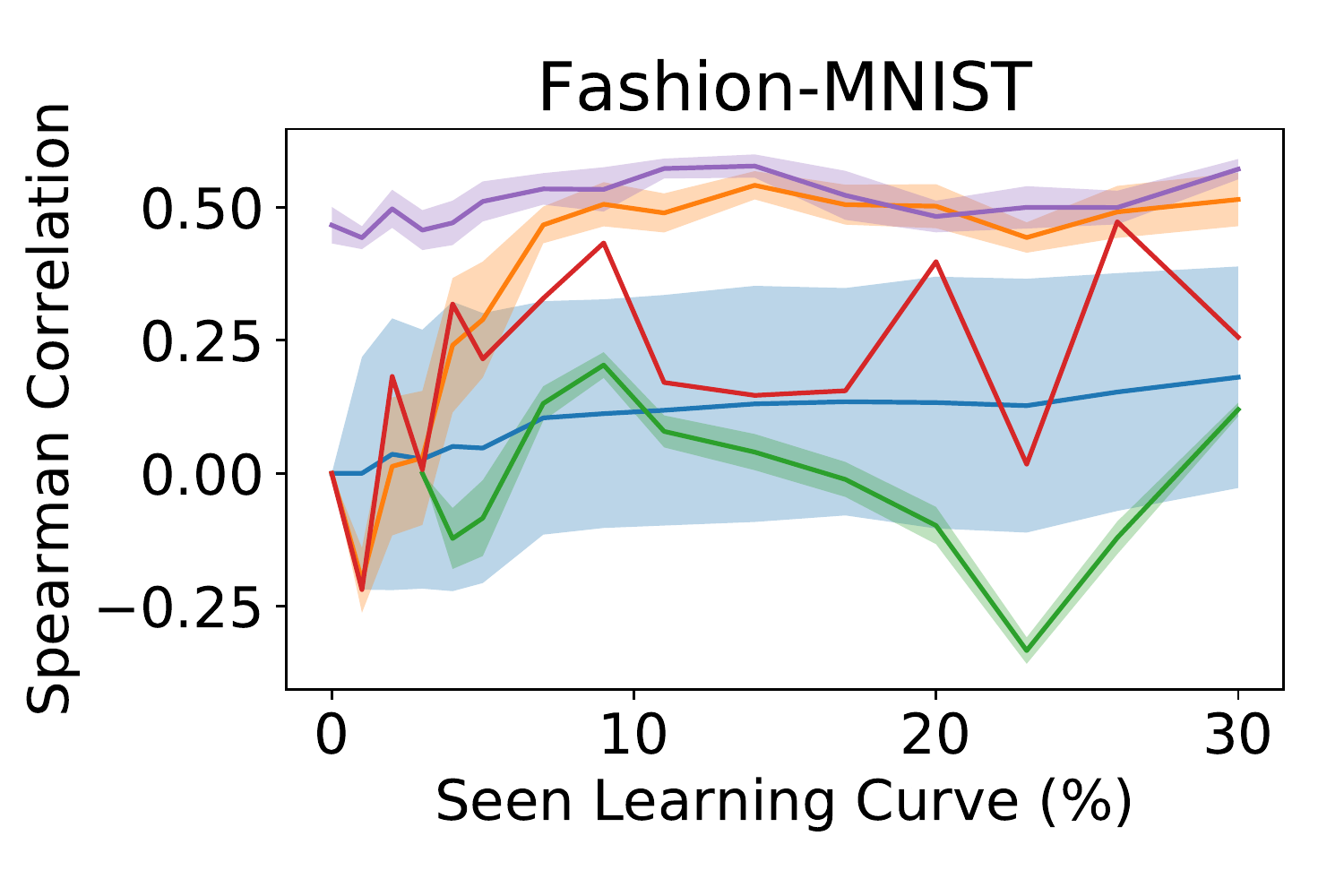}
\includegraphics[width=0.32\textwidth]{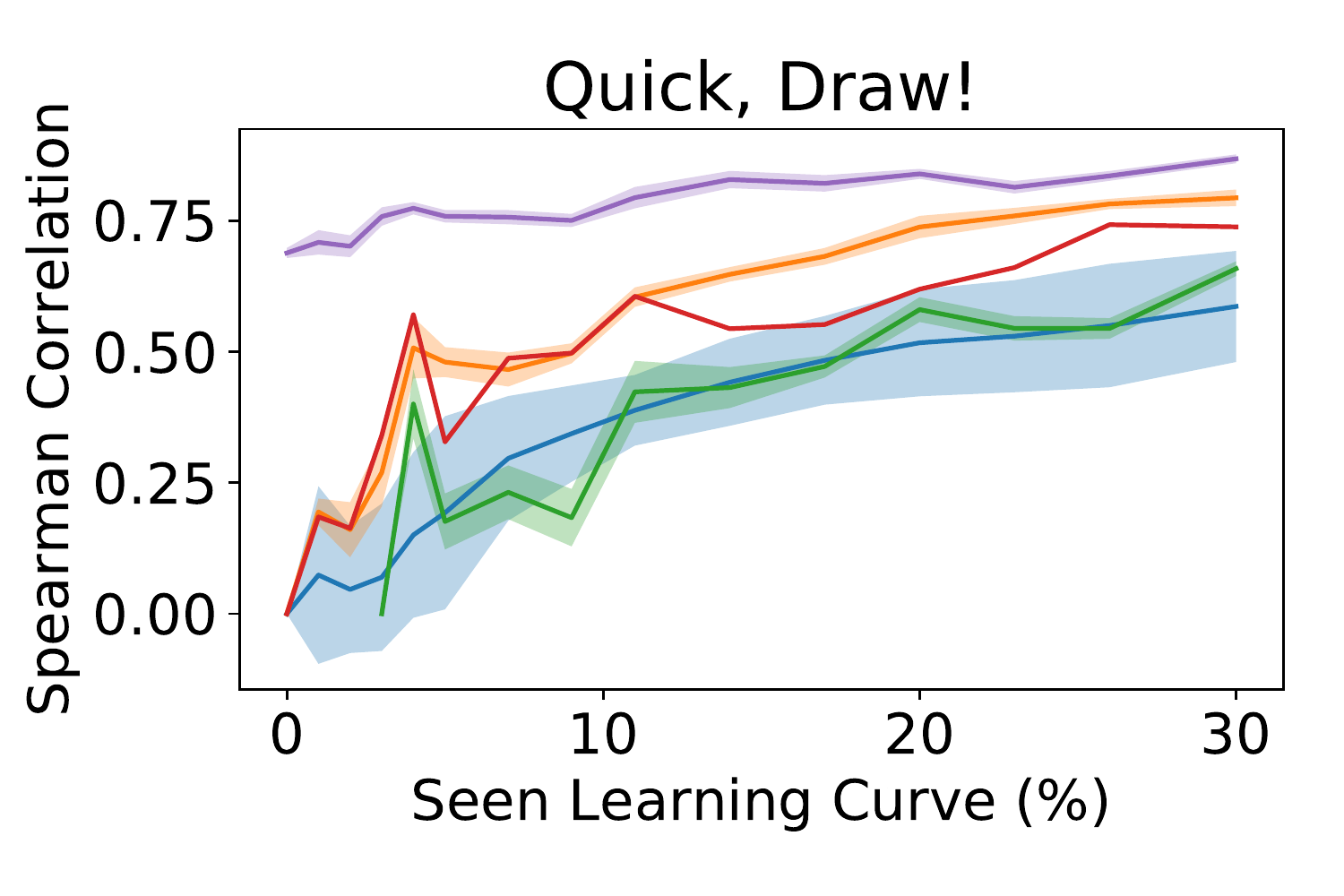}
\includegraphics[width=0.32\textwidth]{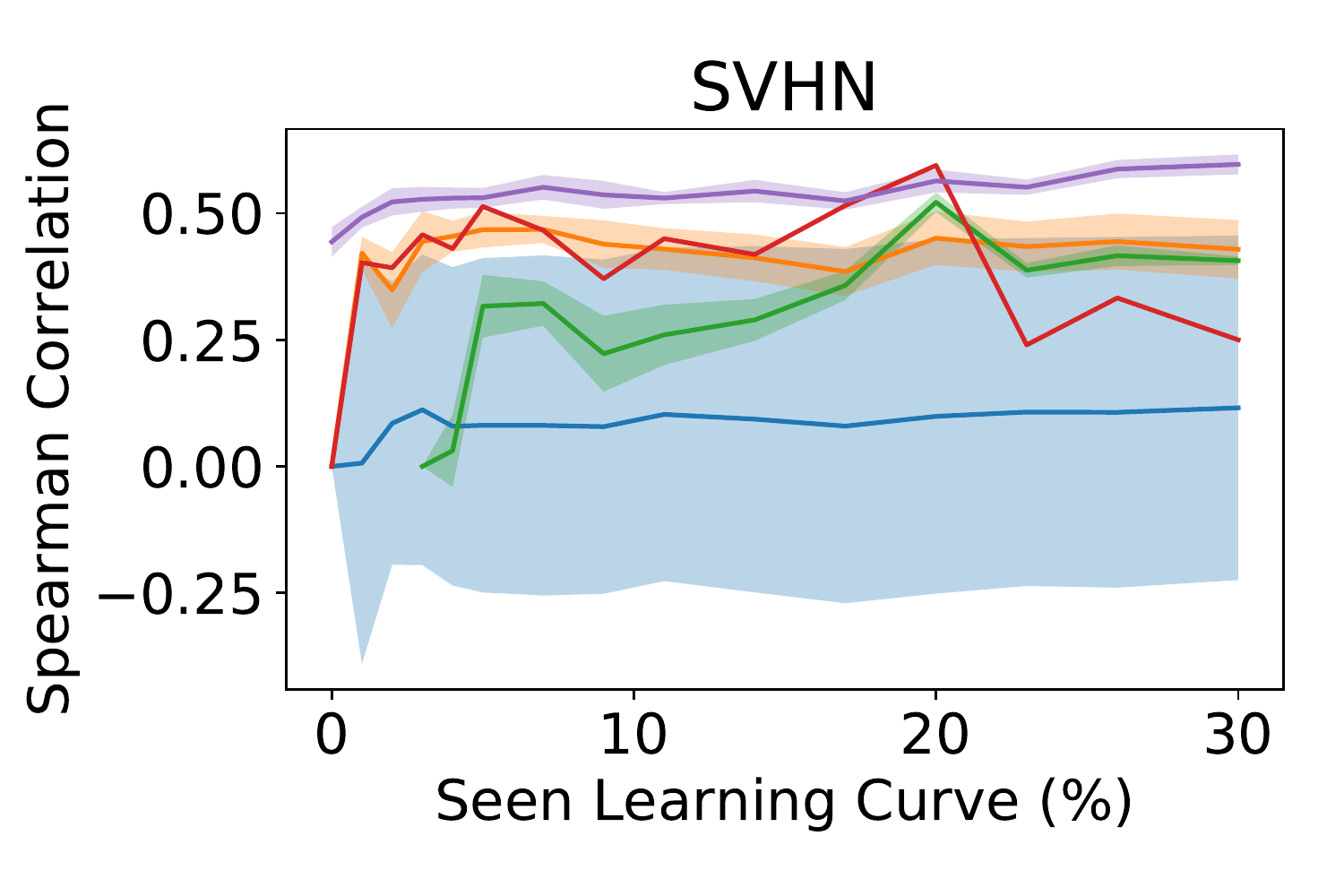}
\includegraphics[trim={14.7cm 2cm 0 2.4cm},clip,width=0.25\textwidth]{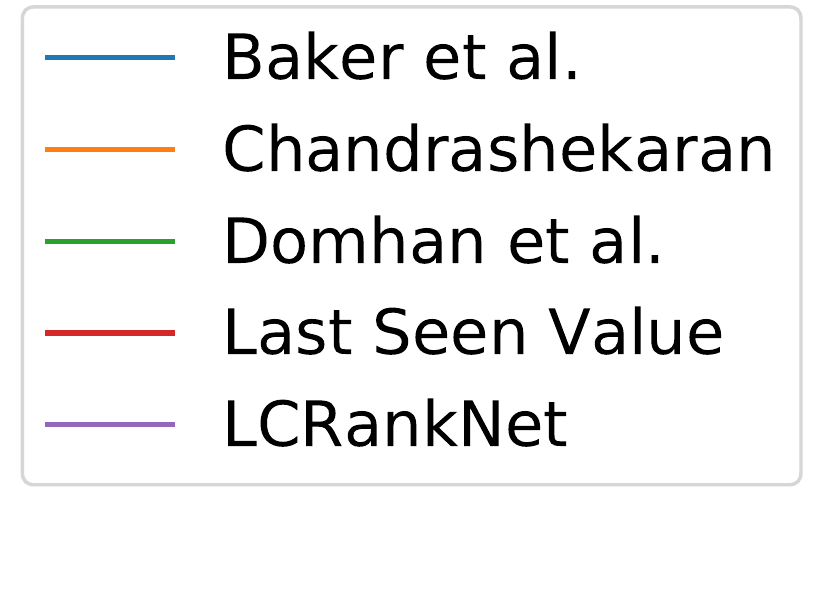}
\caption{The x-axis indicates the length observed of the learning curve.
We report the mean of ten repetitions.
The shaded area is the standard deviation.
Our method \ourLCP{} outperforms its competitors on all datasets.}
\label{fig:lcp-rank-results}
\end{figure*}
\paragraph{Learning curve component}
A CNN is used to process the partial learning curve.
We consider up to four different convolutional layers with different kernel sizes.
The exact number depends on the length of the learning curve since we consider no kernel sizes larger than the learning curve length.
Each convolution is followed by a global max pooling layer and their outputs are concatenated.
This results in the learned representation for the learning curve.
In the appendix we analyze alternative modeling options.

\paragraph{Architecture component}
In this paper we consider two experiments for learning curve prediction.
In one we deal with NAS in the NASNet search space~\citep{Zoph2018}, in the other we work on the joint optimization of architecture and hyperparameters on tabular data~\citep{Klein2019}.
In the former experiment, the search space consists of CNNs, which consist of two cells with five blocks each.
This means that 60 decisions about architecture have to be made.
In the second experiment, the fully connected neural networks (FCNN) consists of only two layers, so the number of decisions is only 6.
For more details refer to the appendix.
We learn an embedding for every option.
An LSTM takes this embedding and generates the architecture embedding.

\paragraph{Dataset component}
As we intend to learn from learning curves of other datasets, we include dataset embeddings as one of the components.
Every dataset has its own embedding and it will be used whenever a learning curve observed on this dataset is selected.
If the dataset is new, then the corresponding embedding is initialized at random.
As the model observes more learning curves from this dataset, it improves the dataset embedding.
The embedding helps us to model dataset-specific peculiarities in order to adjust the ranking if necessary.

\paragraph{Hyperparameter component}
Other hyperparameters, if available, are passed directly to the last layer.

\paragraph{Technical details}
During the development process, we found that the architecture component leads to instabilities during training.
To avoid this, we regularize the output of the LSTM layer by using it as an input to an autoregressive model, similar to a sequence-to-sequence model~\citep{Sutskever2014} that recreates the original description of the architecture.
In addition, we use the attention mechanism~\citep{Bahdanau2014} to facilitate this process.
All parameters of the layers in $f$ are trained jointly by means of Adam~\citep{Kingma2014} by minimizing 
\begin{equation}
    L = \alpha L_{\text{ce}} + \left(1-\alpha\right) L_{\text{rec}}
\end{equation}a weighted linear combination of the ranking loss (Equation \eqref{eq:pairwise-ranking-loss}) and the reconstruction loss with $\alpha=0.8$.
The hyperparameter $\delta$ allows for trading precision vs. recall or search time vs. regret, respectively.
There are two extreme cases: if we set $\delta\geq 1$, every run is terminated immediately.
If we set $\delta\leq 0$, we never terminate any run.
For our experiment we set $\delta=0.45$ which means that if the predicted probability that the new model is better than the best one is below 45\%, the run is terminated early.

\section{Experiments}
In this section, we first discuss how to create the meta-knowledge and then analyze our model in terms of learning curve ranking and the ability to use it as a way to accelerate NAS.
Finally, we examine its individual components and behavior in certain scenarios.
Furthermore, we show that \ourLCP{} can be combined with various optimization methods and used for hyperparameter optimization tasks.

\begin{table*}[t]
\caption{Results obtained by the different methods on five different datasets. For both metrics the smaller, the better. Regret reported in percent, time in GPU hours.}
\label{tab:random-search-results}
\vskip 0.15in
\begin{center}
\begin{small}
\begin{sc}
\begin{tabular}{lrrrrrrrrrr}
\toprule
\bf Method &\multicolumn{2}{c}{\bf CIFAR-10}  &\multicolumn{2}{c}{\bf CIFAR-100}&\multicolumn{2}{c}{\bf Fashion}  &\multicolumn{2}{c}{\bf Quickdraw}&\multicolumn{2}{c}{\bf SVHN}\\
 & Regr. & Time & Regr. & Time & Regr. & Time & Regr. & Time & Regr. & Time\\
\midrule
No Early Termination & 0.00 & 1023 & 0.00 & 1021 & 0.00 & 1218 & 0.00 & 1045 & 0.00 & 1485\\
\citet{Domhan2015} & 0.56 & 346 & 0.82 & 326 & 0.00 & 460 & 0.44 & 331 & 0.28 & 471\\
\citet{Li2017} & 0.22 & 106 & 0.78 & 102 & 0.32 & 132 & 0.54 & 109 & 0.00 & 156\\
\citet{Baker2018} & 0.00 & 89 & 0.00 & 77 & 0.00 & 129 & 0.00 & 107 & 0.00 & 241\\
\citet{Jamieson2016} & 0.62 & 62 & 0.00 & 54 & 0.18 & 70 & 0.40 & 60 & 0.28 & 88\\
\citet{Chandrashekaran2017} & 0.62 & 30 & 0.00 & 35 & 0.28 & 41 & 0.30 & 82 & 0.06 & 164\\
\ourLCP{} & 0.22 & 20 & 0.00 & 11 & 0.10 & 19 & 0.00 & 28 & 0.10 & 74\\
\bottomrule
\end{tabular}
\end{sc}
\end{small}
\end{center}
\vskip -0.1in
\end{table*}
\subsection{Meta-knowledge}
We compare our method to similar methods on five different datasets: CIFAR-10, CIFAR-100, Fashion-MNIST, Quickdraw, and SVHN.
We use the original train/test splits if available.
Quickdraw has a total of 50 million data points and 345 classes.
To reduce the training time, we select a subset of this dataset.
We use 100 different randomly selected classes and choose 300 examples per class for the training split and 100 per class for the test split.
5,000 random data points of the training dataset serve as validation split for all datasets.

To create the meta-knowledge, we choose 200 architectures per dataset at random from the NASNet search space~\citep{Zoph2018} such that we train a total of 1,000 architectures.
We would like to point out that these are 1,000 unique architectures, there is no architecture that has been trained on several different datasets.
Each architecture is trained for 100 epochs with stochastic gradient descent and cosine learning rate schedule without restart~\citep{Loshchilov2017}.

We use standard image preprocessing and augmentation: for every image, we first subtract the channel mean and then divide by the channel standard deviation.
Images are padded by a margin of four pixels and randomly cropped back to the original dimension.
For  all  datasets  but  SVHN  we  apply  random  horizontal flipping.
Additionally, we use Cutout~\citep{Devries2017}.

For the experiments in Section~\ref{sub:hyper-opt} we rely on the tabular benchmark~\citep{Klein2019}.
This benchmark contains meta-data for four regression tasks.
In total 62,208 different FCNN with different architecture and hyperparameters are evaluated.
100 settings per dataset are chosen at random as meta-knowledge.

The following experiments are conducted in a leave-one-dataset-out cross-validation.
That means when considering one dataset, all meta-knowledge but the one for this particular dataset is used.

\subsection{Ranking performance}
First, we analyze the quality of the learning curve rankings by different learning curve prediction methods.
In this experiment we choose 50 different learning curves at random as a test set.
Five random learning curves are used as a training set for every repetition.
Each learning curve prediction method ranks the 50 architectures by observing the partial learning curve whose length varies from 0 to 30 percent of the final learning curve.
We repeat the following experiment ten times and report the mean and standard deviation of the correlation between the true and the predicted ranking in Figure~\ref{fig:lcp-rank-results}.
As a correlation metric, we use Spearman's rank correlation coefficient.
Thus, the correlation is 1 for a perfect ranking and 0 for an uncorrelated, random ranking.
Our method \ourLCP{} shows for all datasets higher correlation.
If there are no or only very short partial learning curves available, our method shows the biggest difference to the existing methods.
The reason for this is a combination of the consideration of the network architecture together with additional meta-knowledge.
We analyze the impact of each component in detail in Section~\ref{sub:ablation-study}.

The method by \citet{Chandrashekaran2017} consistently shows the second best results and in some cases can catch up to the results of our method.
The method by \citet{Baker2018} stands out due to the high standard deviation.
It is by far the method with the smallest squared error on test.
However, the smallest changes in the prediction lead to a significantly different ranking, which explains the high variance in their results.
The method by \citet{Domhan2015} requires a minimum length of the learning curve to make predictions.
Accordingly, we observe rank correlation values starting from a learning curve length of 4\%.
Using the last seen value to determine the ranking of learning curves is a simple yet efficient method~\citep{Klein2017}.
In fact, it is able to outperform some of the more elaborate methods.

\subsection{Accelerating random neural architecture search}
In this experiment, we demonstrate the utility of learning curve predictors in the search for network architectures.
For the sake of simplicity, we accelerate a random search in the NASNet search space~\citep{Zoph2018}.

The random search samples 200 models and trains each of them for 100 epochs to obtain the final accuracy.
In the end, the best of all these models is returned.
Now each learning curve predictor iterates over these sampled architectures in the same order and determines at every third epoch if the training should be aborted.
For Successive Halving~\citep{Jamieson2016} and Hyperband~\citep{Li2017} we follow the algorithm defined by the authors and use the recommended settings.
The current best model discovered after iterating over all 200 architectures is returned as the best model for this learning curve predictor.
The goal of the methods is to minimize the regret compared to a random search without early stopping and at the same time reduce the computational effort.

\begin{figure*}
\centering
\includegraphics[width=0.4\textwidth,valign=c]{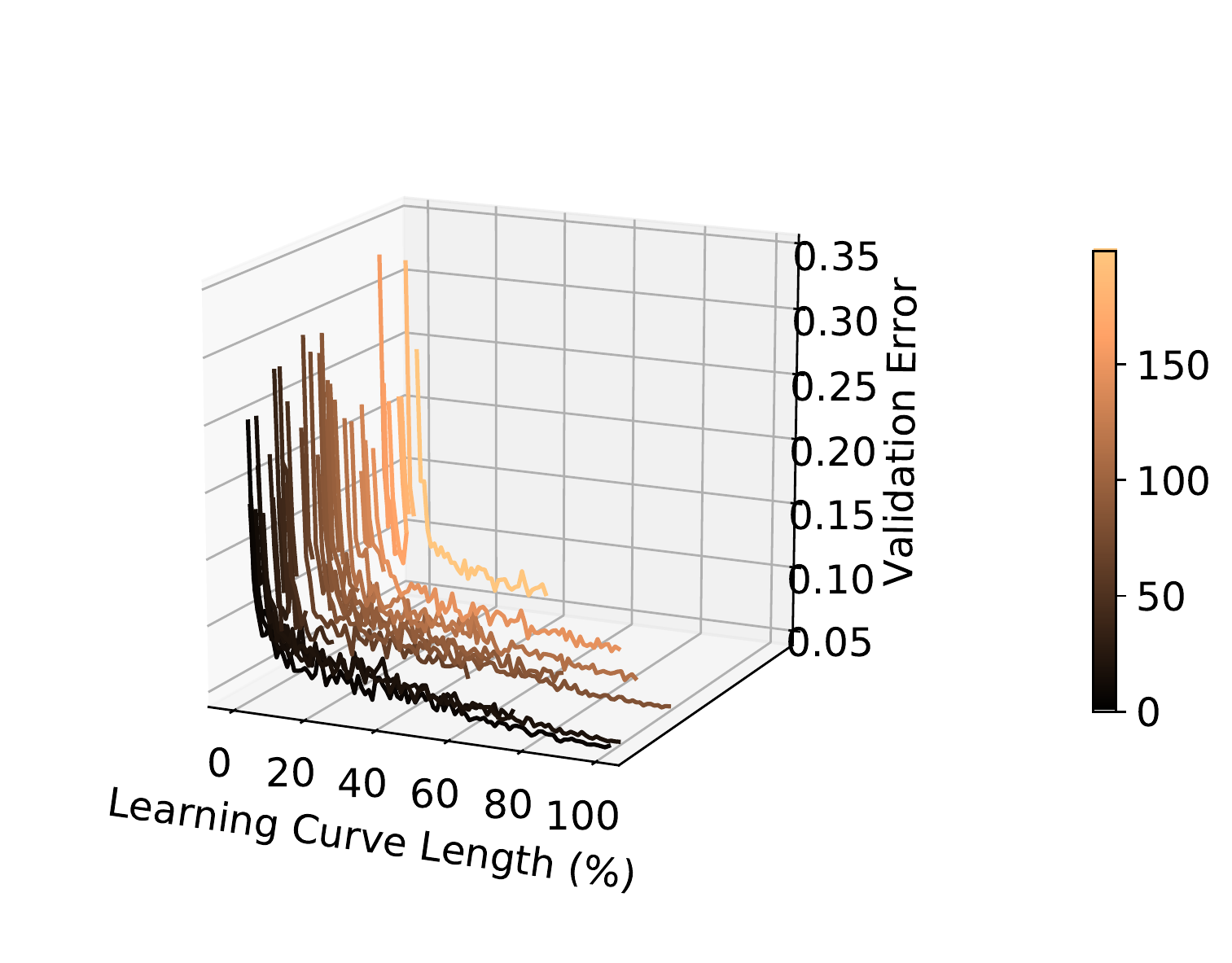}
\includegraphics[width=0.3\textwidth,valign=c]{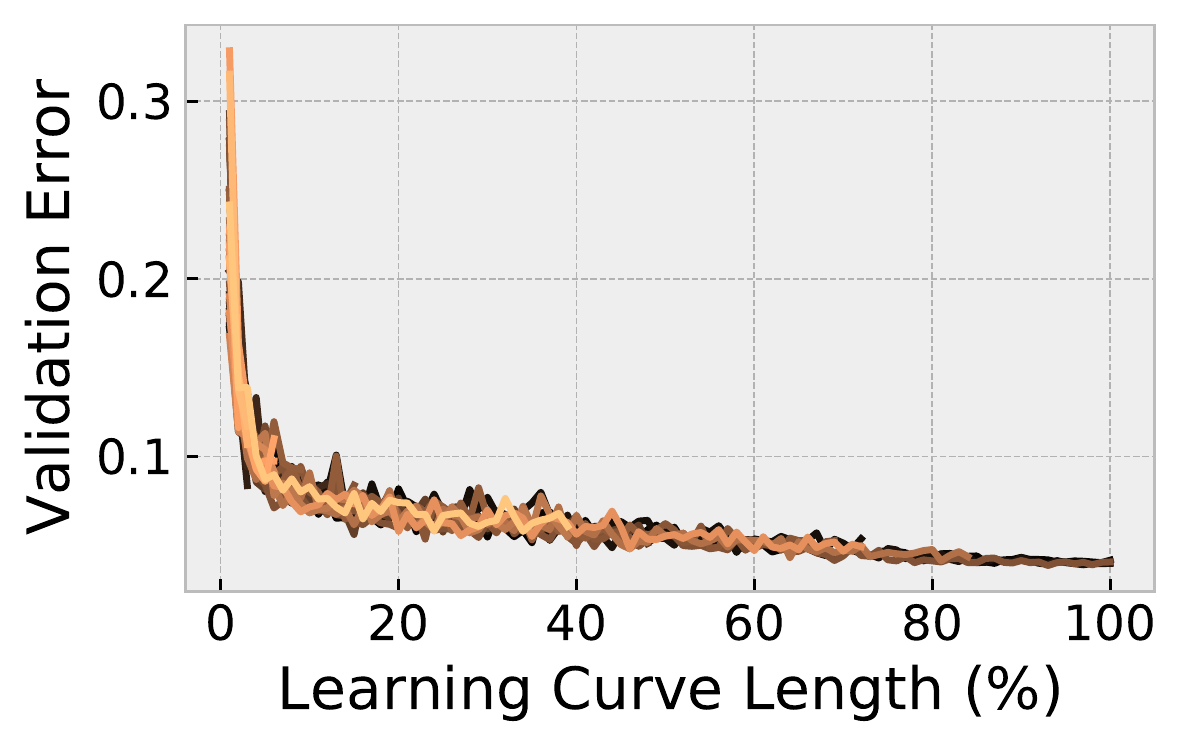}
\caption{\ourLCP{} is speeding up architecture search for SVHN. Difference between learning curves is small, making this the hardest task.}
\label{fig:random-search-experiment}
\end{figure*}
One of our observations is that \citet{Domhan2015}'s method underestimates performance when the learning curve is short.
As a result, the method ends each training process early after only a few epochs.
Therefore, we follow the recommendation by \citet{Domhan2015} and do not end processes before we have seen the first 30 epochs.
We summarize the results in Table~\ref{tab:random-search-results}.
Our first observation is that all methods accelerate the search with little regret.
Here we define regret as the difference between the accuracy of the model found by the random search and the accuracy of the model found by one of the methods.
Not surprisingly, \citet{Domhan2015}'s method takes up most of the time, as it requires significantly longer learning curves to make its decision.
In addition, we can confirm the results by \citet{Baker2018,Chandrashekaran2017}, both of which report better results than \citet{Domhan2015}.
Our method requires the least amount of time for each dataset.
For CIFAR-100 we do not observe any regret, but a reduction of the time by a factor of 100.
In some cases we observe an insignificantly higher regret than some of the other methods.
In our opinion, the time saved makes up for it.

In Figure \ref{fig:random-search-experiment} we visualize the random search for SVHN.
As you can see, many curves not only have similar behavior but also similar error.
For this reason, it is difficult to decide whether to discard a model safely, which explains the increased runtime.
The only methods that do not show this behavior are Successive Halving and Hyperband.
The simple reason is that the number of runs terminated and the time they are discarded are fixed by the algorithm and do not rely on the properties of the learning curve.
The disadvantages are obvious: promising runs may be terminated and unpromising runs may run longer than needed.

Finally, we compare our method with DARTS~\citep{Liu2019}, a method that speeds up the search by parameter sharing.
We train the architecture discovered in the previous experiment with an equivalent training setup like DARTS for a fair comparison.
This architecture achieved a classification error of 2.99\% on CIFAR-10 after only 20 GPU hours were searched.
In comparison, DARTS (1st order) needs 36 GPU hours for a similarly good architecture (3.00\% error).
DARTS (2nd order) can improve these results to 2.76\%, but it takes 96 GPU hours.
One of the major disadvantages of DARTS and related methods is that they cannot generalize to any architecture search space~\citep{Yu2020} and do not allow hyperparameter optimization.
Our proposed method does not suffer from these problems.
\begin{figure*}[t]
\centering
\includegraphics[width=0.3\textwidth]{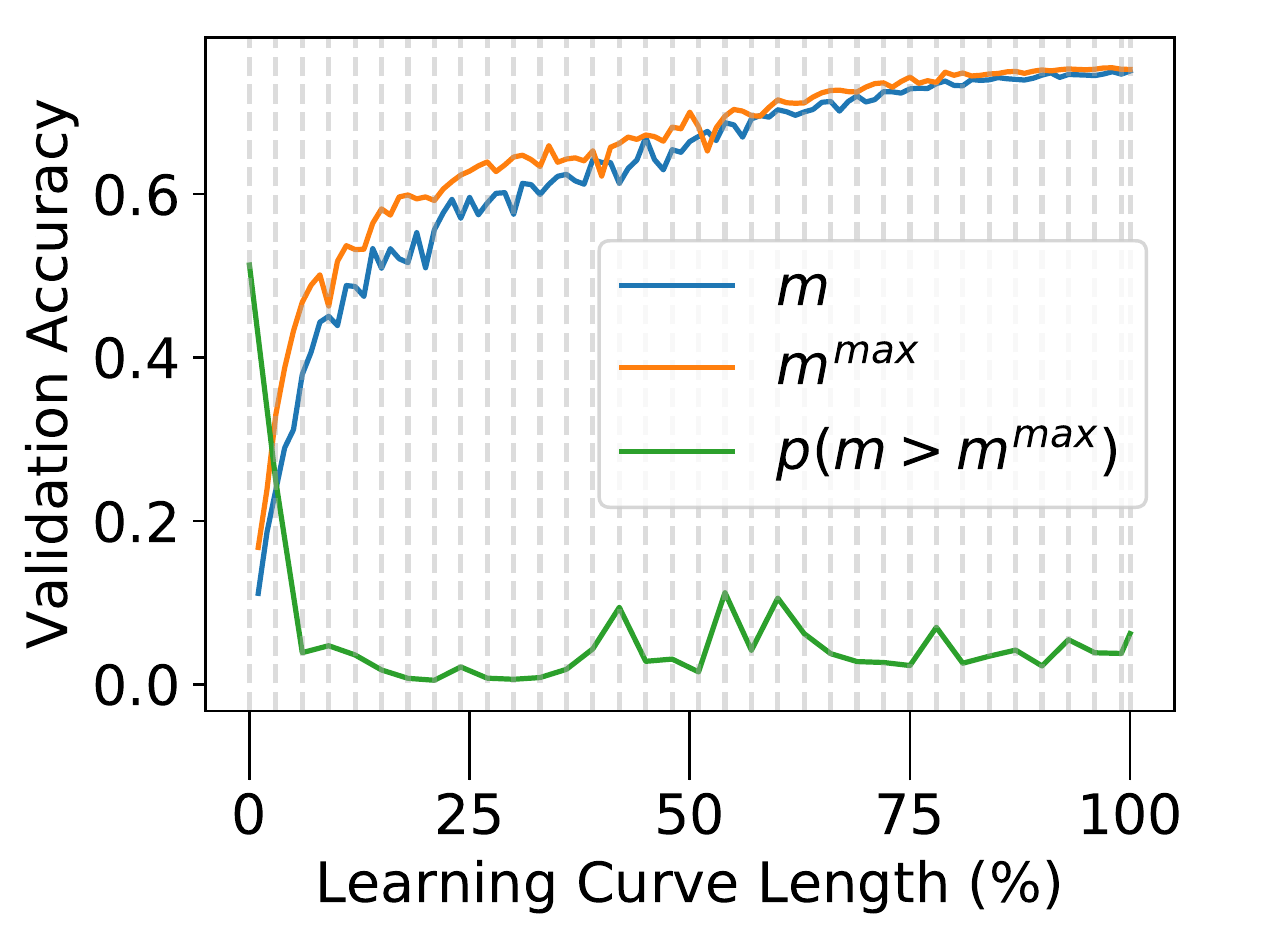}
\includegraphics[width=0.3\textwidth]{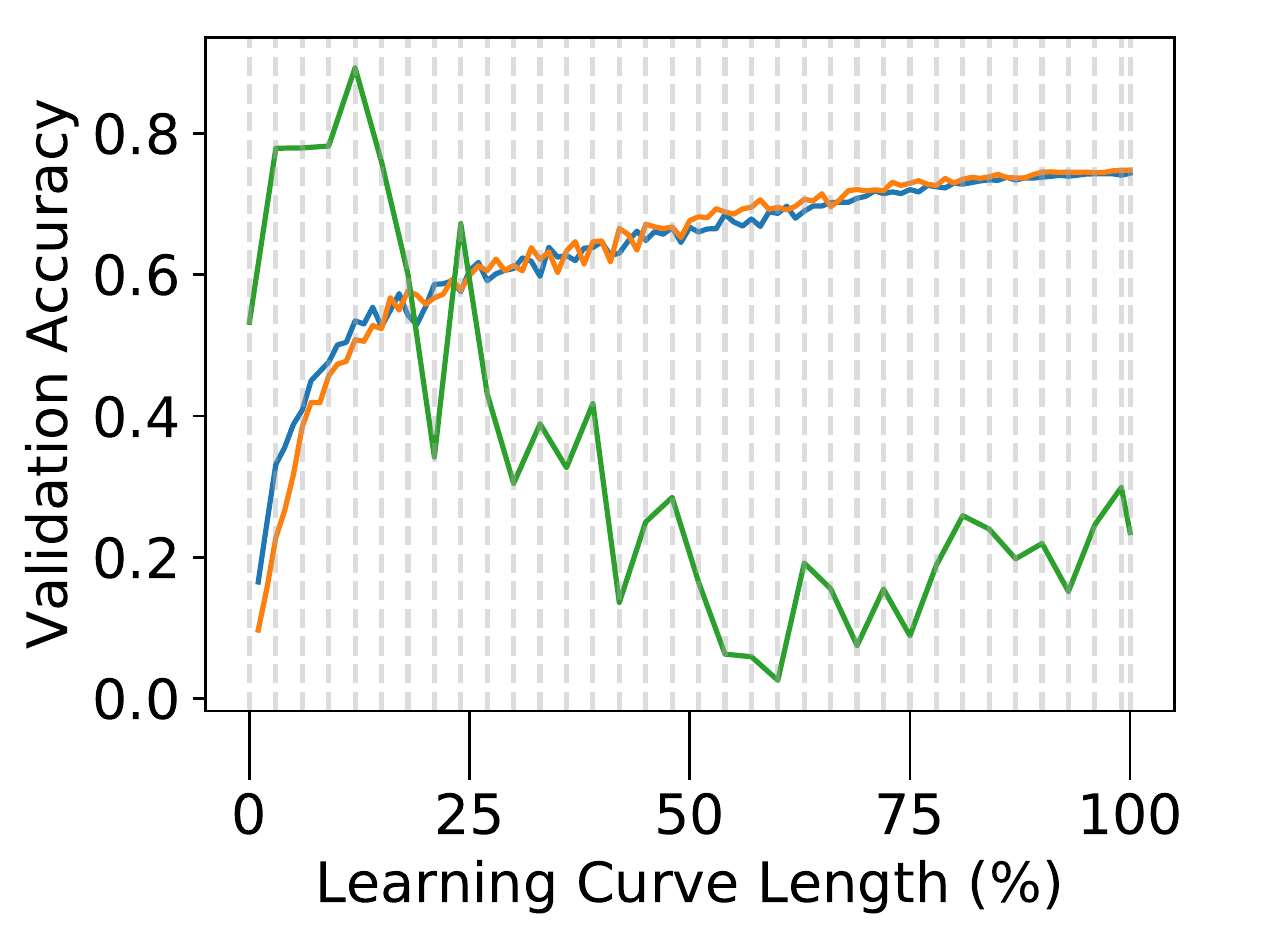}\\
\includegraphics[width=0.3\textwidth]{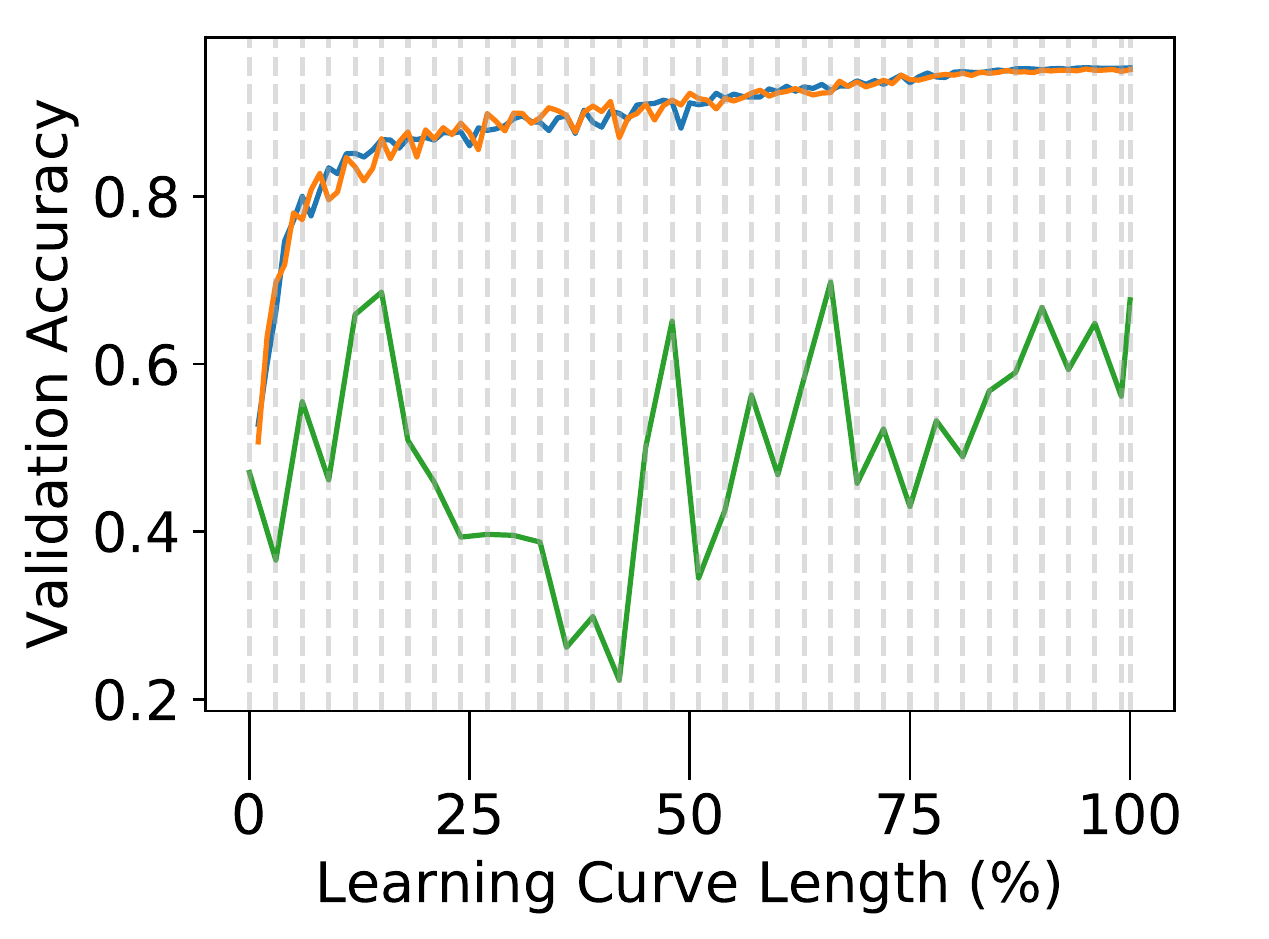}
\includegraphics[width=0.3\textwidth]{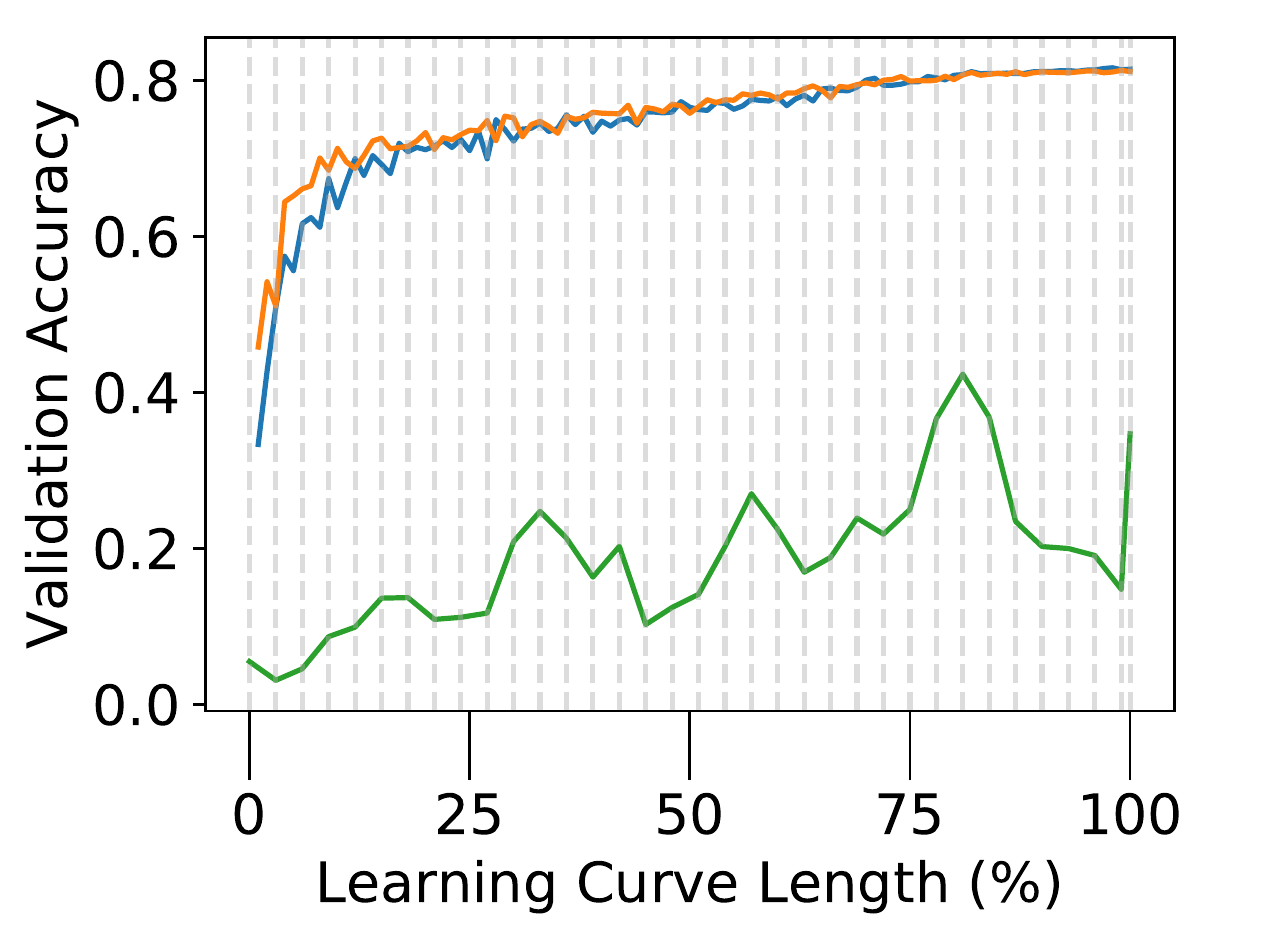}
\caption{Analysis of the predicted probability of \ourLCP{} with growing learning curve length. The plots in the top row show examples for correct decisions by \ourLCP{}. The bottom row shows two examples where its decision was wrong.}
\label{fig:prediction-analysis}
\end{figure*}
\subsection{\ourLCP{} prediction analysis}
We saw in the previous selection that \ourLCP{} does not perform perfectly.
There are cases where its regret is greater than zero or where the search time is higher which indicates that early stopping was applied later than maybe possible.
In this section we try to shed some light on the decisions made by the model and try to give the reader some insight of the model's behavior.

We provide four example decisions of \ourLCP{} in Figure~\ref{fig:prediction-analysis}.
The plots in the top row show cases where \ourLCP{} made a correct decision, the plots in the bottom row are examples for incorrect decisions.

In both of the correct cases (top row), \ourLCP{} assigns a higher probability to begin with, using meta-knowledge.
However, in the top left case it becomes evident after only very few epochs that $m^{\text{max}}$ is consistently better than $m$ such that the probability $p\left(m>m^{\text{max}}\right)$ reduces sharply to values close to zero which would correctly stop training early on.
The case in the top right plot is more tricky.
The probability is increasing until a learning curve length of 12\% as the learning curve seem to indicate that $m$ is better than $m^{\text{max}}$.
However, the learning curve of $m^{\text{max}}$ approaches the values of $m$ and then the probability decreases and training is stopped early.

We continue now with the discussion of the two examples in the bottom row which \ourLCP{} erroneously terminated early, which causes in turn a regret higher than zero in Table~\ref{tab:random-search-results}.
In the bottom left we show an example where sometimes $m$ is better for a longer period and sometimes $m^{\text{max}}$.
This is arguable a very difficult problem and it is hard to predict which curve will eventually be better.
However, \ourLCP{} shows a reasonable behavior.
As the learning curve of $m$ is consistently worse than $m^{\text{max}}$ in the segment from 15\% to 42\%, the probability is decreasing.
Starting from a length of 57\%, where $m$ shows superior performance than $m^{\text{max}}$ for several epochs, the probability starts to raise.
From learning curve length of 70\% onward, both learning curves are very close and the difference in the final accuracy is only 0.0022.
The example visualized in the bottom right plot is a very interesting one.
The learning curve of $m^{\text{max}}$ is consistently better than or almost equal to $m$ up to the very end.
Towards the end, learning curves are getting very close.
In fact, from learning curve length 63\% onward, the maximum difference between $m$ and $m^{\text{max}}$ per epoch is only 0.008.
Hence, in the beginning it seems like a trivial decision to reject $m$.
However, eventually $m$ turns out to be better than $m^{\text{max}}$.

\begin{figure*}[t]
\centering
\includegraphics[width=0.32\textwidth]{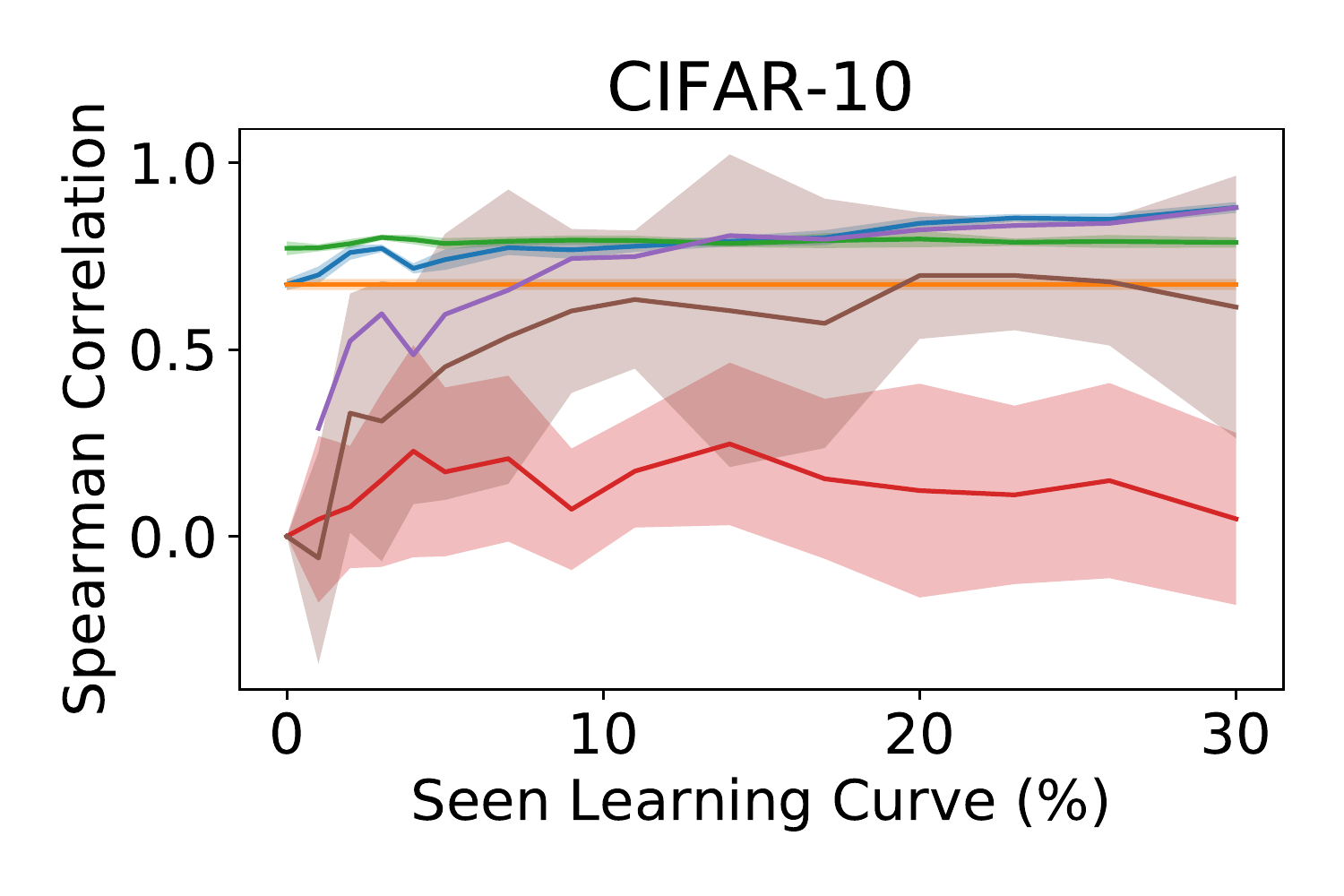}
\includegraphics[width=0.32\textwidth]{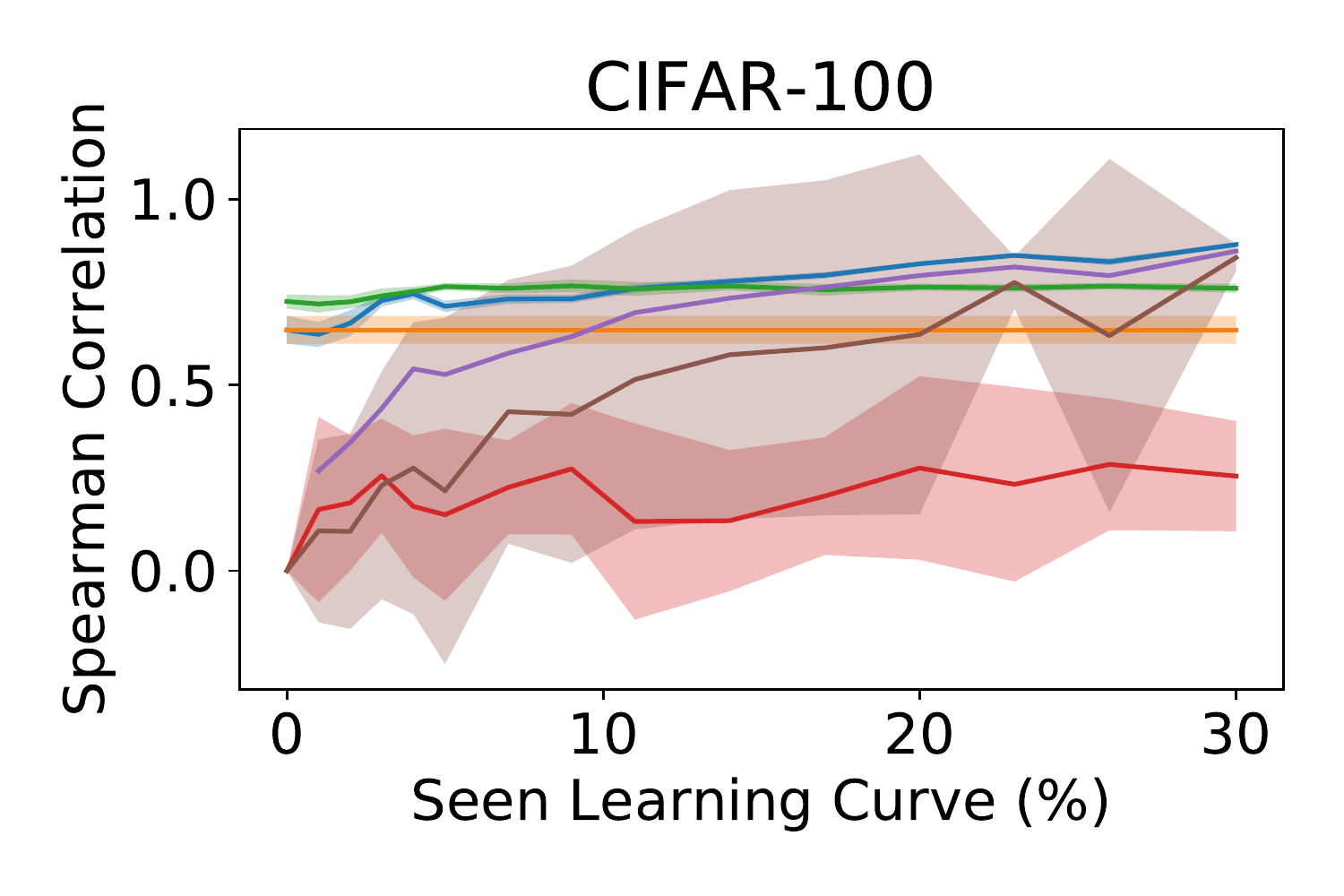}
\includegraphics[width=0.32\textwidth]{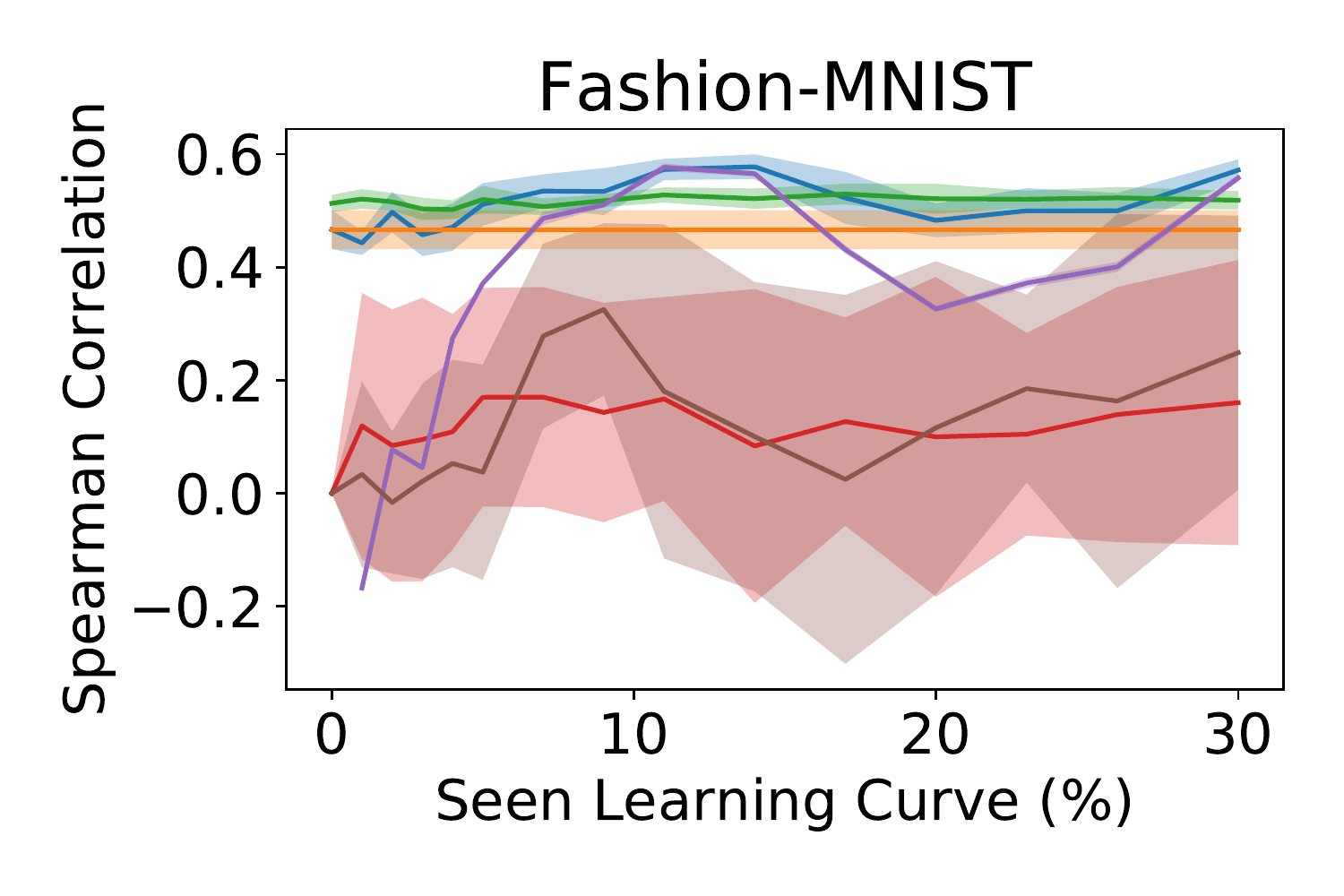}
\includegraphics[width=0.32\textwidth]{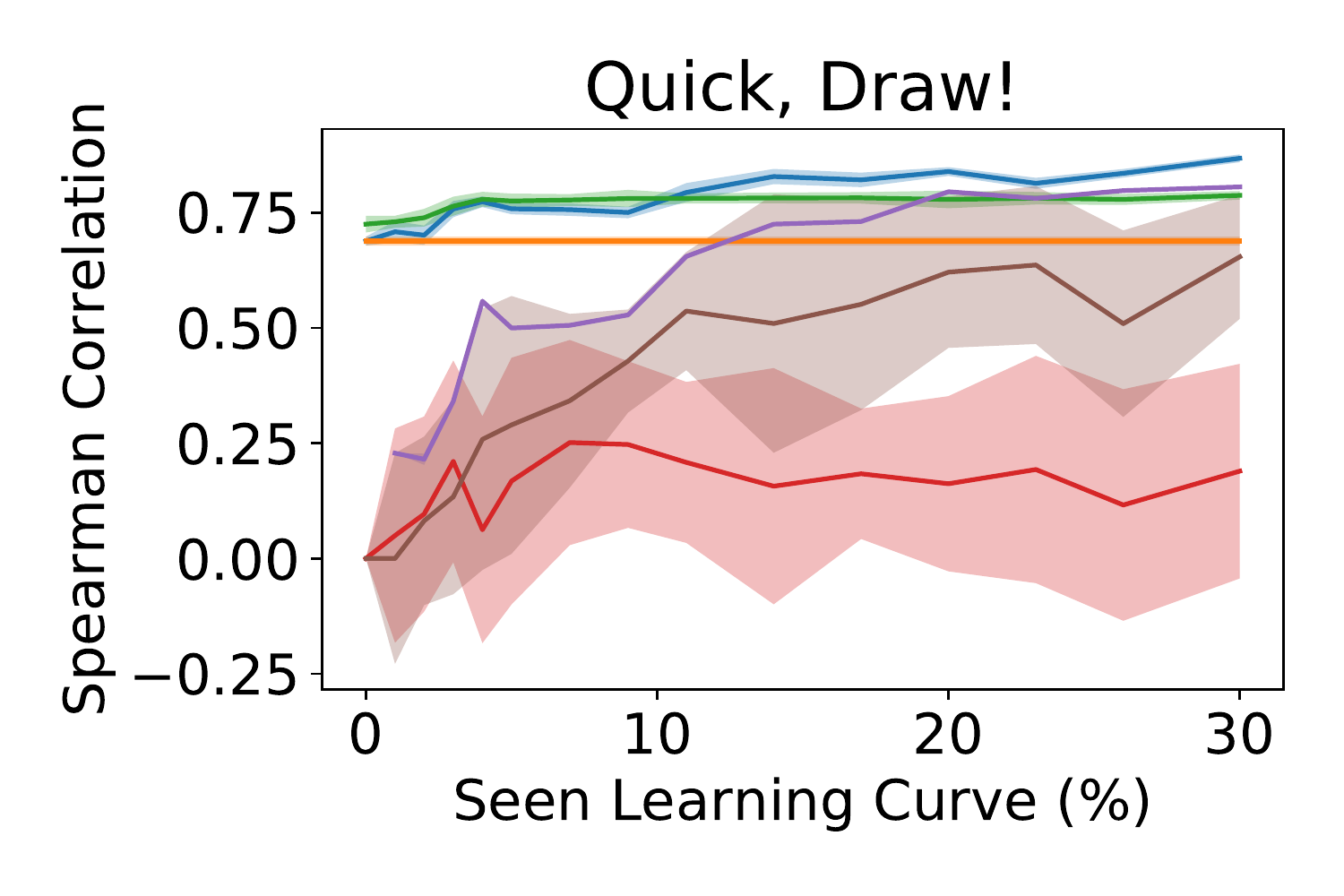}
\includegraphics[width=0.32\textwidth]{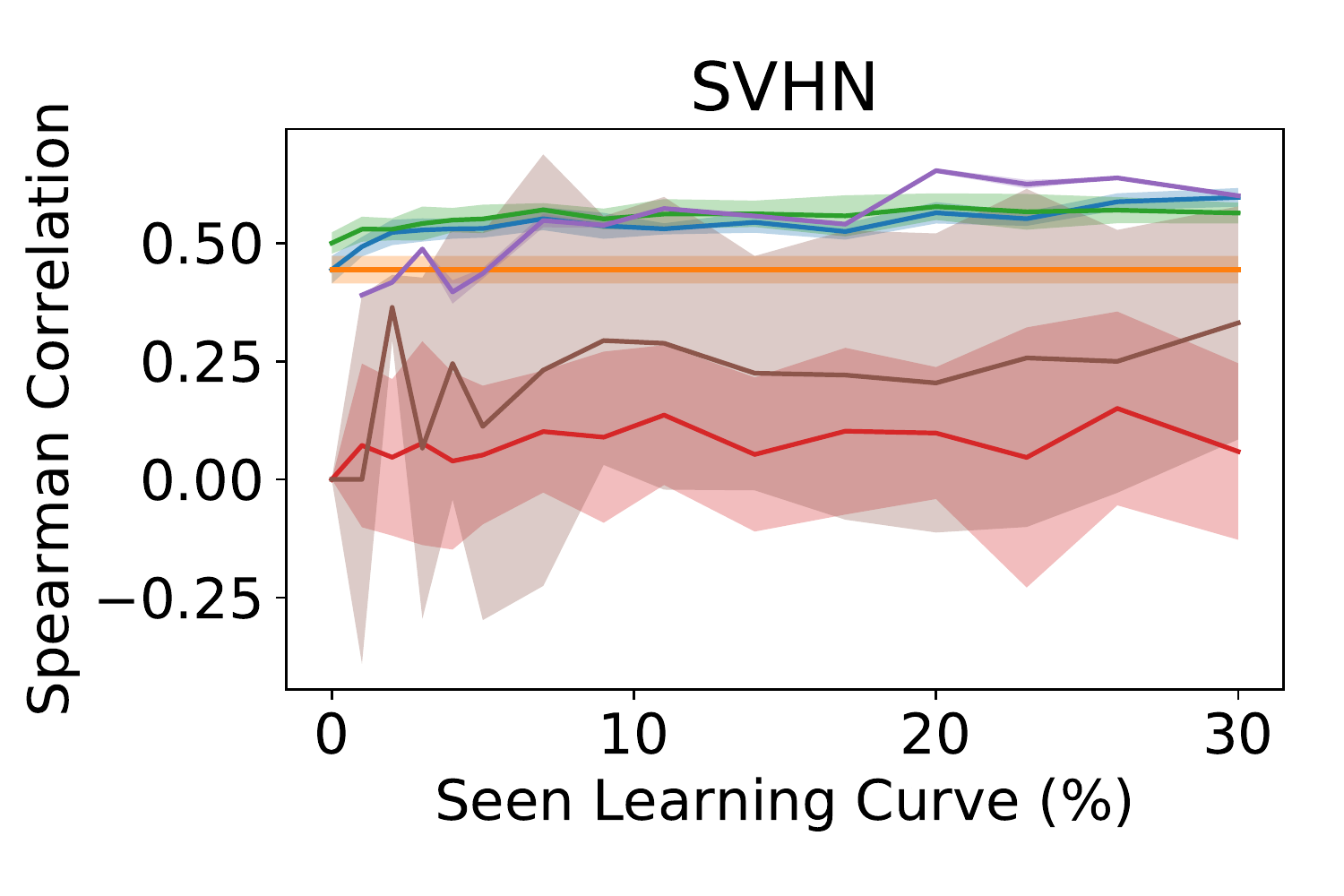}
\includegraphics[trim={14.7cm 0 0 0},clip,width=0.32\textwidth]{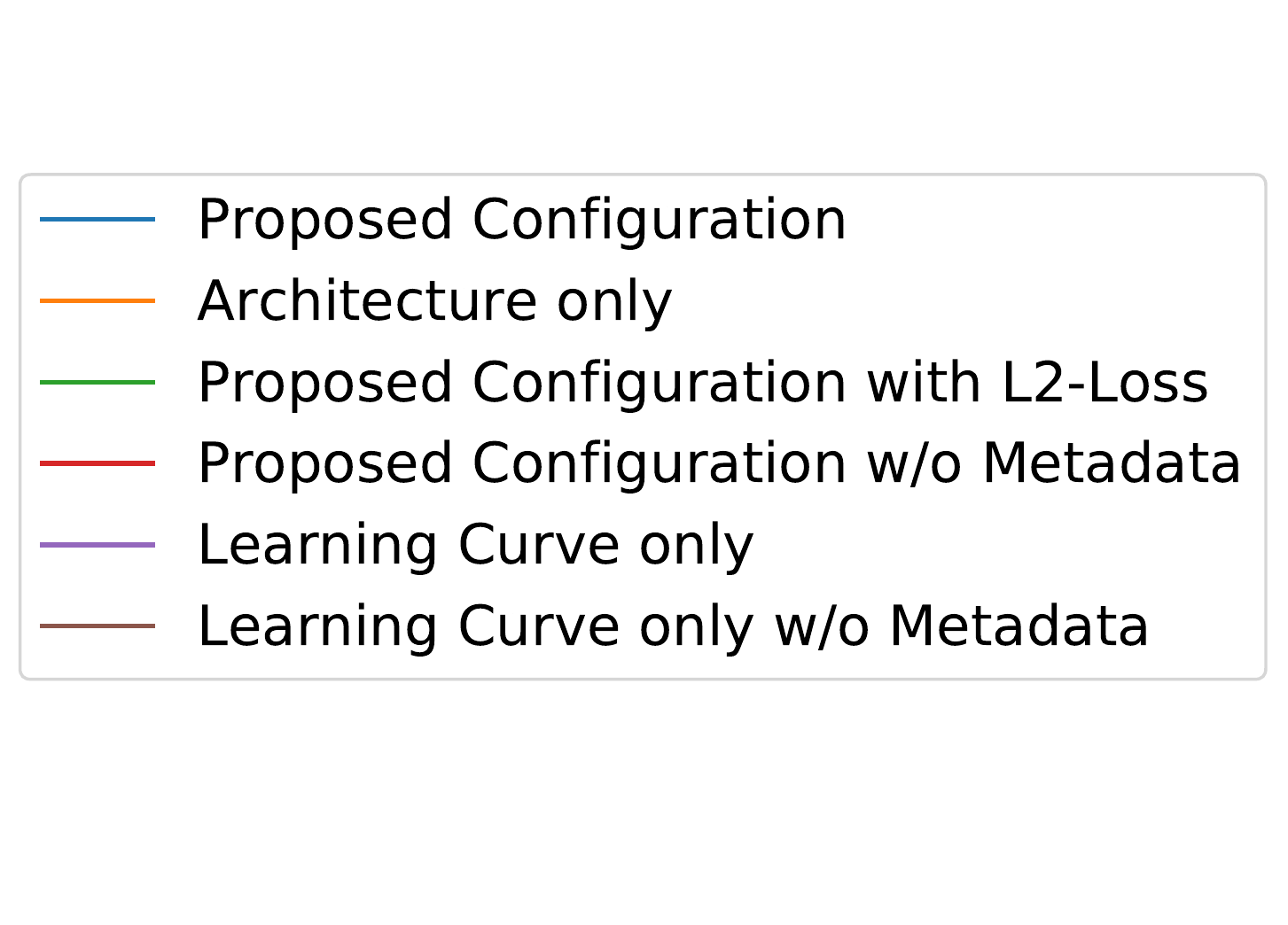}
\caption{Analysis of the different components of \ourLCP{}. Every single component, metadata, consideration of the learning curve and architecture description, is vital.}
\label{fig:lcp-rank-results-ablation}
\end{figure*}
In conclusion, deciding whether one model will be better than another one based on a partial learning curve is a challenging task.
A model that turns out to be (slightly) better than another one can be dominated consistently for most of the learning curve.
This makes it a very challenging problem for automated methods and human experts.

\subsection{Analysis of \ourLCP{}'s components}\label{sub:ablation-study}
We briefly mentioned before which components of our learning curve ranker have an essential influence on its quality.
We would like to deepen the analysis at this point and compare the configuration we have proposed with different variants in Figure \ref{fig:lcp-rank-results-ablation}.
We consider variants with and without metadata, architecture description or learning curve consideration.
In addition, we compare our configuration trained with pairwise ranking loss to one trained with a pointwise ranking loss.

One of the most striking observations is that the metadata is essential for the model.
This is not surprising since in particular the learning of architecture embedding needs sufficient data.
Sufficient data is not available in this setup, so we observe a much higher variance for these variants.
Even the variant that only considers the learning curve benefits from additional meta-knowledge.
But even this is not surprising, since stagnating learning processes show similar behavior regardless of the dataset.
Using the meta-knowledge, both components achieve good results on their own.
It can be clearly seen that these components are orthogonal to one another.
The variant, which only considers the architecture, shows very good results for short learning curves.
If only the learning curve is considered, the results are not very good at first, but improve significantly with the length of the learning curve.
A combination of both methods ultimately leads to our method and further improves the results.
Finally, we compare the use of a pointwise ranking loss (L2 loss) versus a pairwise ranking loss.
Although our assumption was that the latter would have to be fundamentally better, since it optimizes the model parameters directly for the task at hand, in practice this does not necessarily seem to be the case.
Especially for short learning curves, the simpler method achieves better results.
However, once the learning curve is long enough, the pairwise ranking loss pays off.

\subsection{Accelerating hyperparameter optimization}\label{sub:hyper-opt}
\ourLCP{} is not limited to NAS, but can be used for hyperparameter optimization methods as well.
We demonstrate this by combining \ourLCP{} with TPE~\citep{Bergstra2011}, Regularized Evolution (RE)~\citep{Real2019}), and Reinforcement Learning (RL)~\citep{Zoph2017} to optimize the hyperparameters of an FCNN using a tabular data benchmark~\citep{Klein2019}.
The objective of this benchmark is to minimize the mean squared error (MSE) by choosing the right architecture and hyperparameter settings.
All experiments are repeated ten times.
Each optimization method can evaluate up to 100 different configurations and chooses the best one based on its validation MSE.
We report results with respect to test regret following \citet{Klein2019}.
This number indicates the difference between the test MSE actually achieved and the best possible test MSE.
In our experiments, we observed that early termination generally improves the method (Figure~\ref{fig:early-stop-protein}, Appendix).
All other results are listed in the Appendix.

\begin{figure}[t]
\centering
\includegraphics[width=0.7\columnwidth]{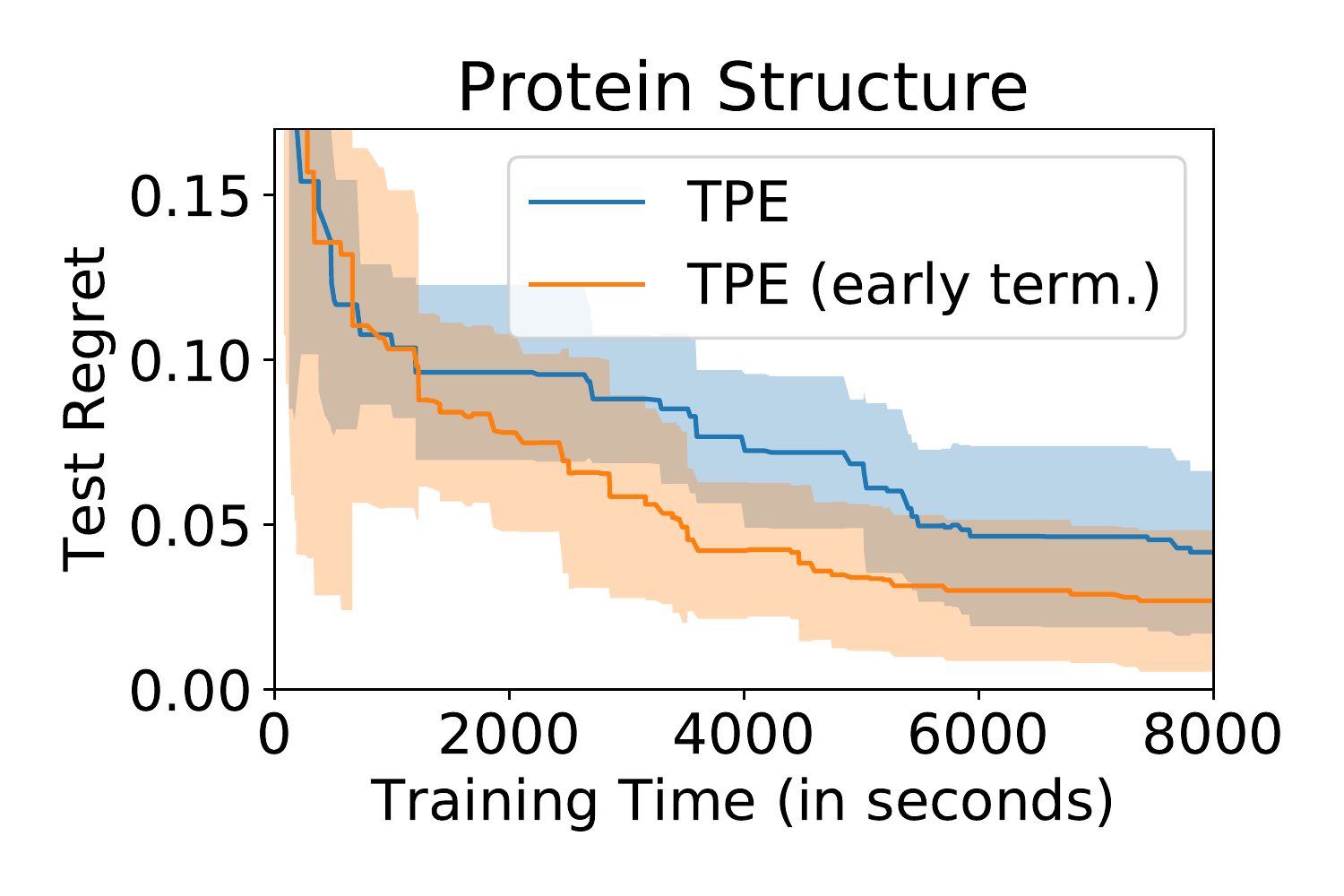}
\includegraphics[width=0.7\columnwidth]{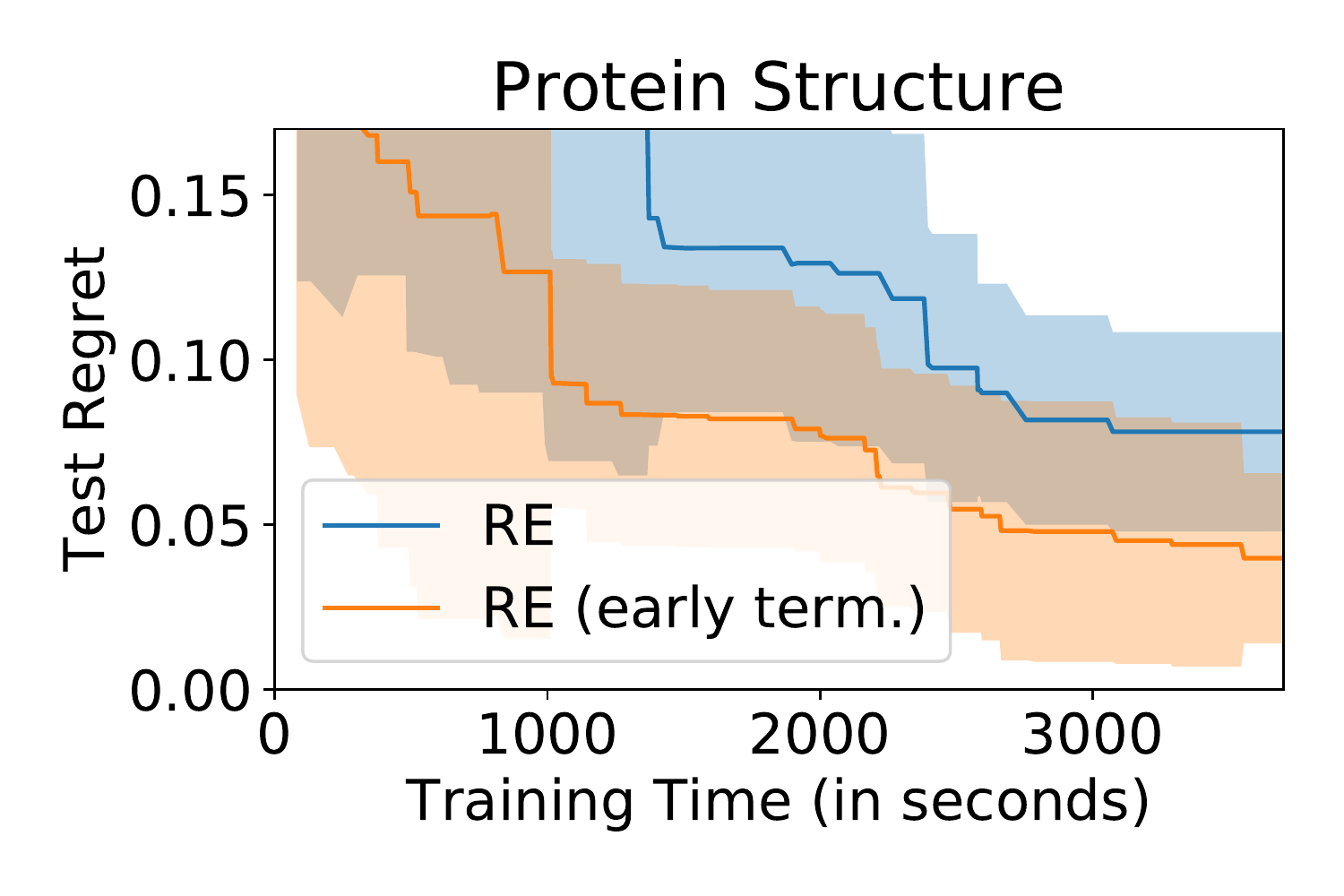}
\caption{TPE and RE can leverage \ourLCP{} to accelerate optimization.
Results for reinforcement learning, more datasets and further insights can be found in the Appendix.}
\label{fig:early-stop-protein}
\end{figure}
\section{Conclusion}
In this paper we present \ourLCP{}, a method to automatically terminate unpromising model configurations early.
The two main novelties of the underlying model are that it is able to consider learning curves from other datasets and that it uses a pairwise ranking loss.
The former allows to predict for relatively short, and in extreme cases even without, learning curves.
The latter directly allows to model the probability that one configuration is better than the another.
We analyze our method on five different datasets against three alternatives.
In an experiment to optimize network architectures, we obtain the fastest results.
In the best case, \ourLCP{} is 100 times faster without sacrificing accuracy.
We also examine the components and predictions of our method to give the reader a better understanding of the design choices and functionalities.
Finally, we demonstrate its use in combination with various hyperparameter optimizers.
\bibliography{main}

\begin{thebibliography}{24}
\providecommand{\natexlab}[1]{#1}
\providecommand{\url}[1]{\texttt{#1}}
\expandafter\ifx\csname urlstyle\endcsname\relax
  \providecommand{\doi}[1]{doi: #1}\else
  \providecommand{\doi}{doi: \begingroup \urlstyle{rm}\Url}\fi

\bibitem[Bahdanau et~al.(2015)Bahdanau, Cho, and Bengio]{Bahdanau2014}
Bahdanau, D., Cho, K., and Bengio, Y.
\newblock Neural machine translation by jointly learning to align and
  translate.
\newblock In \emph{3rd International Conference on Learning Representations,
  {ICLR} 2015, San Diego, CA, USA, May 7-9, 2015, Conference Track
  Proceedings}, 2015.
\newblock URL \url{http://arxiv.org/abs/1409.0473}.

\bibitem[Baker et~al.(2018)Baker, Gupta, Raskar, and Naik]{Baker2018}
Baker, B., Gupta, O., Raskar, R., and Naik, N.
\newblock Accelerating neural architecture search using performance prediction.
\newblock In \emph{6th International Conference on Learning Representations,
  {ICLR} 2018, Vancouver, BC, Canada, April 30 - May 3, 2018, Workshop Track
  Proceedings}, 2018.
\newblock URL \url{https://openreview.net/forum?id=HJqk3N1vG}.

\bibitem[Bergstra et~al.(2011)Bergstra, Bardenet, Bengio, and
  K{\'{e}}gl]{Bergstra2011}
Bergstra, J., Bardenet, R., Bengio, Y., and K{\'{e}}gl, B.
\newblock Algorithms for hyper-parameter optimization.
\newblock In \emph{Advances in Neural Information Processing Systems 24: 25th
  Annual Conference on Neural Information Processing Systems 2011. Proceedings
  of a meeting held 12-14 December 2011, Granada, Spain}, pp.\  2546--2554,
  2011.
\newblock URL
  \url{http://papers.nips.cc/paper/4443-algorithms-for-hyper-parameter-optimization}.

\bibitem[Burges et~al.(2005)Burges, Shaked, Renshaw, Lazier, Deeds, Hamilton,
  and Hullender]{Burges2005}
Burges, C. J.~C., Shaked, T., Renshaw, E., Lazier, A., Deeds, M., Hamilton, N.,
  and Hullender, G.~N.
\newblock Learning to rank using gradient descent.
\newblock In \emph{Machine Learning, Proceedings of the Twenty-Second
  International Conference {(ICML} 2005), Bonn, Germany, August 7-11, 2005},
  pp.\  89--96, 2005.
\newblock \doi{10.1145/1102351.1102363}.
\newblock URL \url{https://doi.org/10.1145/1102351.1102363}.

\bibitem[Chandrashekaran \& Lane(2017)Chandrashekaran and
  Lane]{Chandrashekaran2017}
Chandrashekaran, A. and Lane, I.~R.
\newblock Speeding up hyper-parameter optimization by extrapolation of learning
  curves using previous builds.
\newblock In \emph{Machine Learning and Knowledge Discovery in Databases -
  European Conference, {ECML} {PKDD} 2017, Skopje, Macedonia, September 18-22,
  2017, Proceedings, Part {I}}, volume 10534 of \emph{Lecture Notes in Computer
  Science}, pp.\  477--492. Springer, 2017.

\bibitem[DeVries \& Taylor(2017)DeVries and Taylor]{Devries2017}
DeVries, T. and Taylor, G.~W.
\newblock Improved regularization of convolutional neural networks with cutout.
\newblock \emph{CoRR}, abs/1708.04552, 2017.
\newblock URL \url{http://arxiv.org/abs/1708.04552}.

\bibitem[Domhan et~al.(2015)Domhan, Springenberg, and Hutter]{Domhan2015}
Domhan, T., Springenberg, J.~T., and Hutter, F.
\newblock Speeding up automatic hyperparameter optimization of deep neural
  networks by extrapolation of learning curves.
\newblock In \emph{Proceedings of the Twenty-Fourth International Joint
  Conference on Artificial Intelligence, {IJCAI} 2015, Buenos Aires, Argentina,
  July 25-31, 2015}, pp.\  3460--3468. {AAAI} Press, 2015.

\bibitem[Falkner et~al.(2018)Falkner, Klein, and Hutter]{Falkner2018}
Falkner, S., Klein, A., and Hutter, F.
\newblock {BOHB:} robust and efficient hyperparameter optimization at scale.
\newblock In \emph{Proceedings of the 35th International Conference on Machine
  Learning, {ICML} 2018, Stockholmsm{\"{a}}ssan, Stockholm, Sweden, July 10-15,
  2018}, pp.\  1436--1445, 2018.
\newblock URL \url{http://proceedings.mlr.press/v80/falkner18a.html}.

\bibitem[Jamieson \& Talwalkar(2016)Jamieson and Talwalkar]{Jamieson2016}
Jamieson, K.~G. and Talwalkar, A.
\newblock Non-stochastic best arm identification and hyperparameter
  optimization.
\newblock In \emph{Proceedings of the 19th International Conference on
  Artificial Intelligence and Statistics, {AISTATS} 2016, Cadiz, Spain, May
  9-11, 2016}, pp.\  240--248, 2016.
\newblock URL \url{http://proceedings.mlr.press/v51/jamieson16.html}.

\bibitem[Kingma \& Ba(2015)Kingma and Ba]{Kingma2014}
Kingma, D.~P. and Ba, J.
\newblock Adam: {A} method for stochastic optimization.
\newblock In \emph{3rd International Conference on Learning Representations,
  {ICLR} 2015, San Diego, CA, USA, May 7-9, 2015, Conference Track
  Proceedings}, 2015.
\newblock URL \url{http://arxiv.org/abs/1412.6980}.

\bibitem[Klein \& Hutter(2019)Klein and Hutter]{Klein2019}
Klein, A. and Hutter, F.
\newblock Tabular benchmarks for joint architecture and hyperparameter
  optimization.
\newblock \emph{CoRR}, abs/1905.04970, 2019.
\newblock URL \url{http://arxiv.org/abs/1905.04970}.

\bibitem[Klein et~al.(2017)Klein, Falkner, Springenberg, and Hutter]{Klein2017}
Klein, A., Falkner, S., Springenberg, J.~T., and Hutter, F.
\newblock Learning curve prediction with bayesian neural networks.
\newblock In \emph{5th International Conference on Learning Representations,
  {ICLR} 2017, Toulon, France, April 24-26, 2017, Conference Track
  Proceedings}, 2017.
\newblock URL \url{https://openreview.net/forum?id=S11KBYclx}.

\bibitem[Li et~al.(2017)Li, Jamieson, DeSalvo, Rostamizadeh, and
  Talwalkar]{Li2017}
Li, L., Jamieson, K.~G., DeSalvo, G., Rostamizadeh, A., and Talwalkar, A.
\newblock Hyperband: {A} novel bandit-based approach to hyperparameter
  optimization.
\newblock \emph{J. Mach. Learn. Res.}, 18:\penalty0 185:1--185:52, 2017.
\newblock URL \url{http://jmlr.org/papers/v18/16-558.html}.

\bibitem[Liu et~al.(2019)Liu, Simonyan, and Yang]{Liu2019}
Liu, H., Simonyan, K., and Yang, Y.
\newblock {DARTS:} differentiable architecture search.
\newblock In \emph{7th International Conference on Learning Representations,
  {ICLR} 2019, New Orleans, LA, USA, May 6-9, 2019}, 2019.
\newblock URL \url{https://openreview.net/forum?id=S1eYHoC5FX}.

\bibitem[Liu(2011)]{Liu2011}
Liu, T.
\newblock \emph{Learning to Rank for Information Retrieval}.
\newblock Springer, 2011.
\newblock ISBN 978-3-642-14266-6.
\newblock \doi{10.1007/978-3-642-14267-3}.
\newblock URL \url{https://doi.org/10.1007/978-3-642-14267-3}.

\bibitem[Loshchilov \& Hutter(2017)Loshchilov and Hutter]{Loshchilov2017}
Loshchilov, I. and Hutter, F.
\newblock {SGDR:} stochastic gradient descent with warm restarts.
\newblock In \emph{5th International Conference on Learning Representations,
  {ICLR} 2017, Toulon, France, April 24-26, 2017, Conference Track
  Proceedings}, 2017.
\newblock URL \url{https://openreview.net/forum?id=Skq89Scxx}.

\bibitem[Luo et~al.(2018)Luo, Tian, Qin, Chen, and Liu]{Luo2018}
Luo, R., Tian, F., Qin, T., Chen, E., and Liu, T.
\newblock Neural architecture optimization.
\newblock In \emph{Advances in Neural Information Processing Systems 31: Annual
  Conference on Neural Information Processing Systems 2018, NeurIPS 2018, 3-8
  December 2018, Montr{\'{e}}al, Canada}, pp.\  7827--7838, 2018.
\newblock URL
  \url{http://papers.nips.cc/paper/8007-neural-architecture-optimization}.

\bibitem[Pham et~al.(2018)Pham, Guan, Zoph, Le, and Dean]{Pham2018}
Pham, H., Guan, M.~Y., Zoph, B., Le, Q.~V., and Dean, J.
\newblock Efficient neural architecture search via parameter sharing.
\newblock In \emph{Proceedings of the 35th International Conference on Machine
  Learning, {ICML} 2018, Stockholmsm{\"{a}}ssan, Stockholm, Sweden, July 10-15,
  2018}, pp.\  4092--4101, 2018.
\newblock URL \url{http://proceedings.mlr.press/v80/pham18a.html}.

\bibitem[Real et~al.(2019)Real, Aggarwal, Huang, and Le]{Real2019}
Real, E., Aggarwal, A., Huang, Y., and Le, Q.~V.
\newblock Regularized evolution for image classifier architecture search.
\newblock In \emph{The Thirty-Third {AAAI} Conference on Artificial
  Intelligence, {AAAI} 2019, The Thirty-First Innovative Applications of
  Artificial Intelligence Conference, {IAAI} 2019, The Ninth {AAAI} Symposium
  on Educational Advances in Artificial Intelligence, {EAAI} 2019, Honolulu,
  Hawaii, USA, January 27 - February 1, 2019}, pp.\  4780--4789, 2019.
\newblock \doi{10.1609/aaai.v33i01.33014780}.
\newblock URL \url{https://doi.org/10.1609/aaai.v33i01.33014780}.

\bibitem[Sutskever et~al.(2014)Sutskever, Vinyals, and Le]{Sutskever2014}
Sutskever, I., Vinyals, O., and Le, Q.~V.
\newblock Sequence to sequence learning with neural networks.
\newblock In \emph{Advances in Neural Information Processing Systems 27: Annual
  Conference on Neural Information Processing Systems 2014, December 8-13 2014,
  Montreal, Quebec, Canada}, pp.\  3104--3112, 2014.
\newblock URL
  \url{http://papers.nips.cc/paper/5346-sequence-to-sequence-learning-with-neural-networks}.

\bibitem[Wistuba et~al.(2019)Wistuba, Rawat, and Pedapati]{Wistuba2019}
Wistuba, M., Rawat, A., and Pedapati, T.
\newblock A survey on neural architecture search.
\newblock \emph{CoRR}, abs/1905.01392, 2019.
\newblock URL \url{http://arxiv.org/abs/1905.01392}.

\bibitem[Yu et~al.(2020)Yu, Sciuto, Jaggi, Musat, and Salzmann]{Yu2020}
Yu, K., Sciuto, C., Jaggi, M., Musat, C., and Salzmann, M.
\newblock Evaluating the search phase of neural architecture search.
\newblock In \emph{8th International Conference on Learning Representations,
  {ICLR} 2020, Addis Ababa, Ethiopia, April 26-30, 2020, Conference Track
  Proceedings}, 2020.
\newblock URL \url{https://openreview.net/forum?id=H1loF2NFwr}.

\bibitem[Zoph \& Le(2017)Zoph and Le]{Zoph2017}
Zoph, B. and Le, Q.~V.
\newblock Neural architecture search with reinforcement learning.
\newblock In \emph{5th International Conference on Learning Representations,
  {ICLR} 2017, Toulon, France, April 24-26, 2017, Conference Track
  Proceedings}, 2017.
\newblock URL \url{https://openreview.net/forum?id=r1Ue8Hcxg}.

\bibitem[Zoph et~al.(2018)Zoph, Vasudevan, Shlens, and Le]{Zoph2018}
Zoph, B., Vasudevan, V., Shlens, J., and Le, Q.~V.
\newblock Learning transferable architectures for scalable image recognition.
\newblock In \emph{2018 {IEEE} Conference on Computer Vision and Pattern
  Recognition, {CVPR} 2018, Salt Lake City, UT, USA, June 18-22, 2018}, pp.\
  8697--8710, 2018.
\newblock \doi{10.1109/CVPR.2018.00907}.
\newblock URL
  \url{http://openaccess.thecvf.com/content\_cvpr\_2018/html/Zoph\_Learning\_Transferable\_Architectures\_CVPR\_2018\_paper.html}.

\end{thebibliography}
\bibliographystyle{icml2020}

\clearpage
\appendix
\begin{figure*}
\centering
\includegraphics[width=0.32\textwidth]{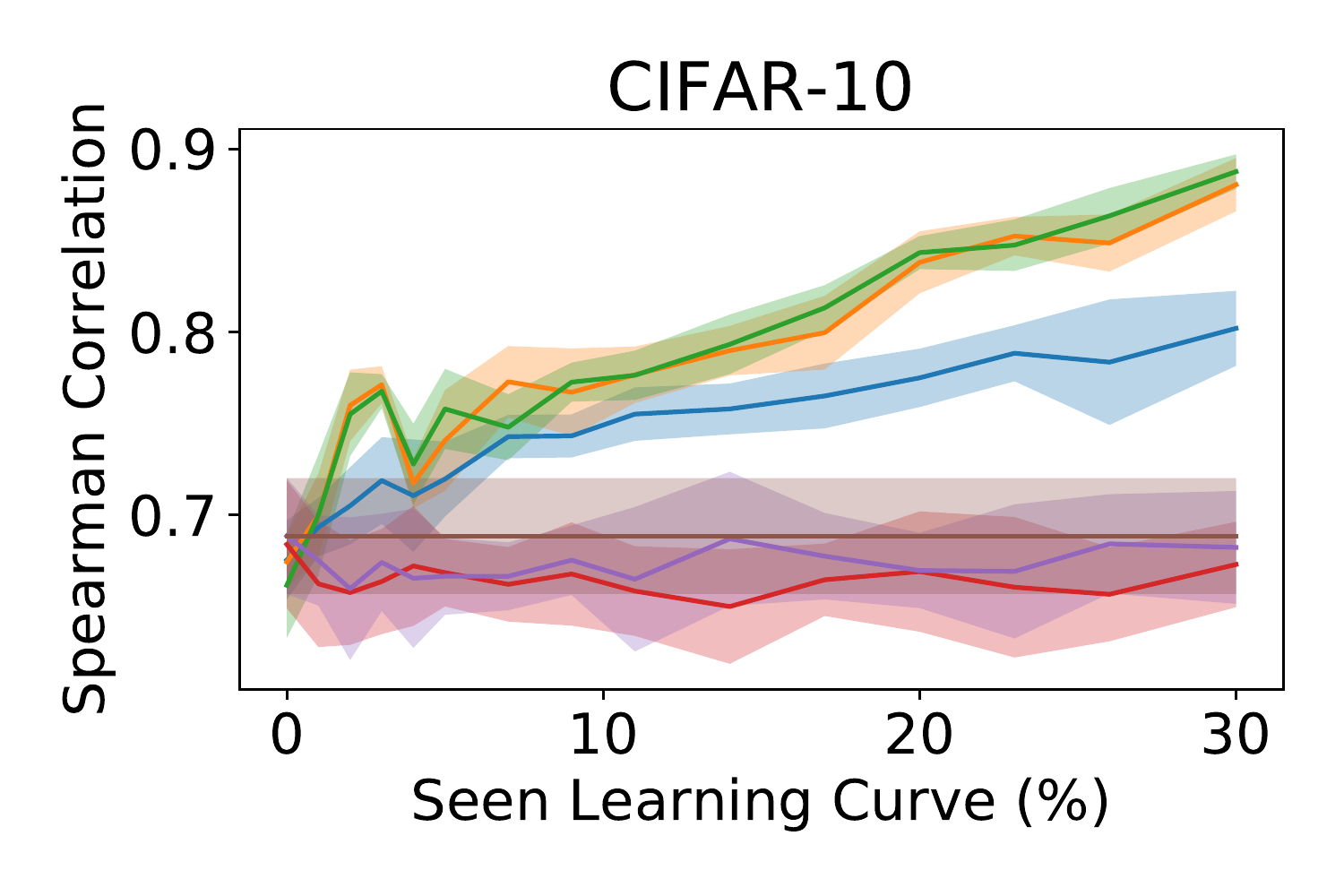}
\includegraphics[width=0.32\textwidth]{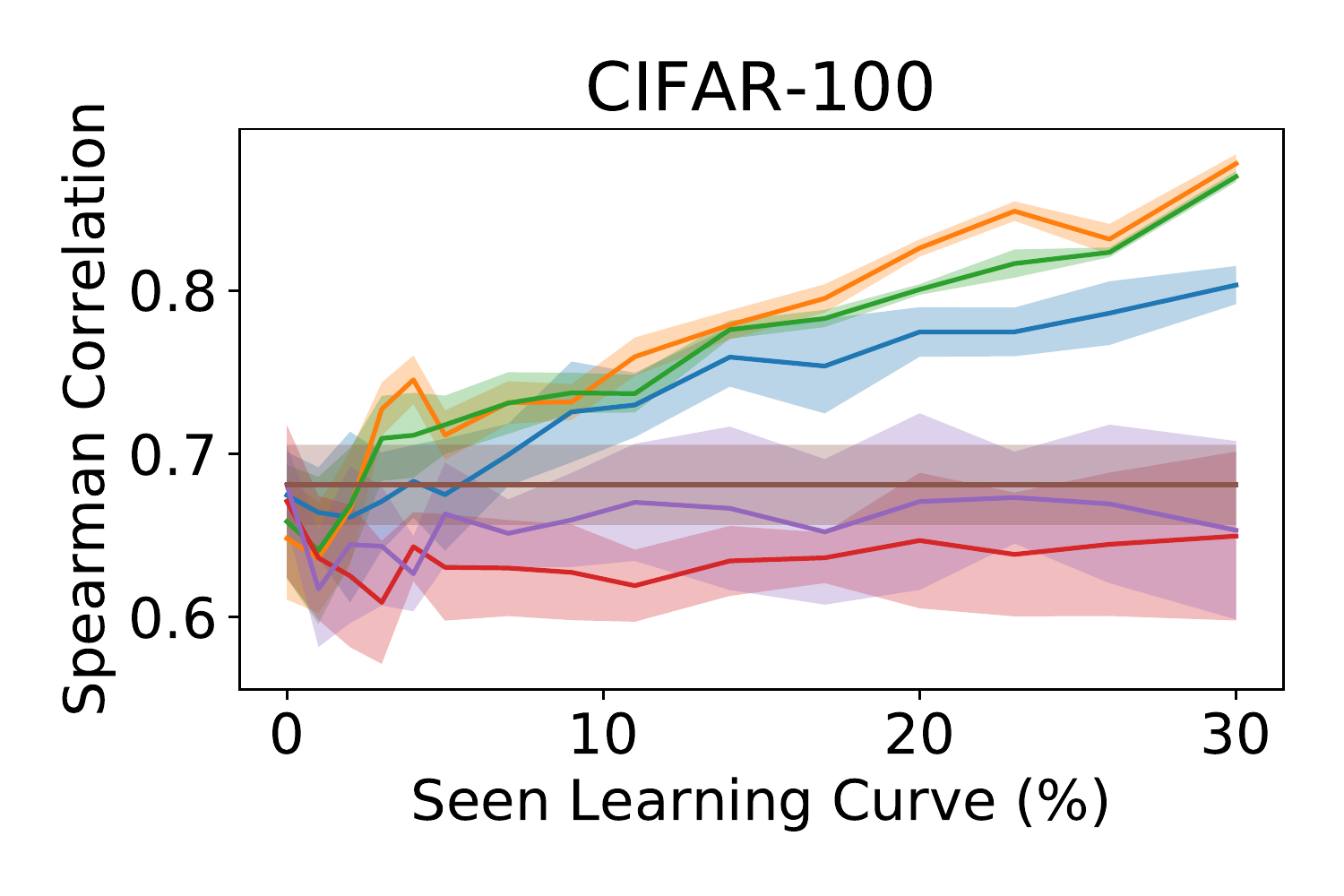}
\includegraphics[width=0.32\textwidth]{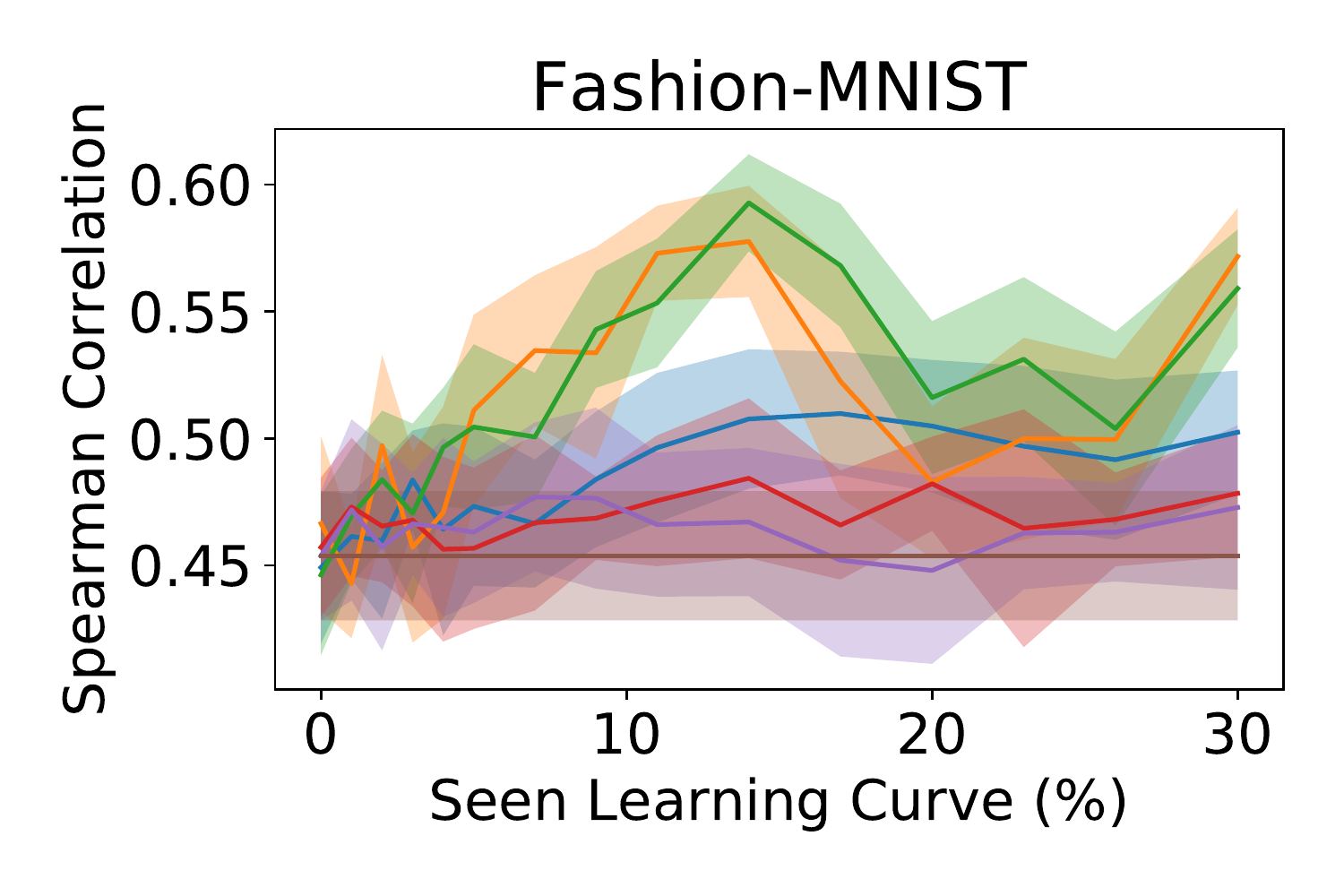}
\includegraphics[width=0.32\textwidth]{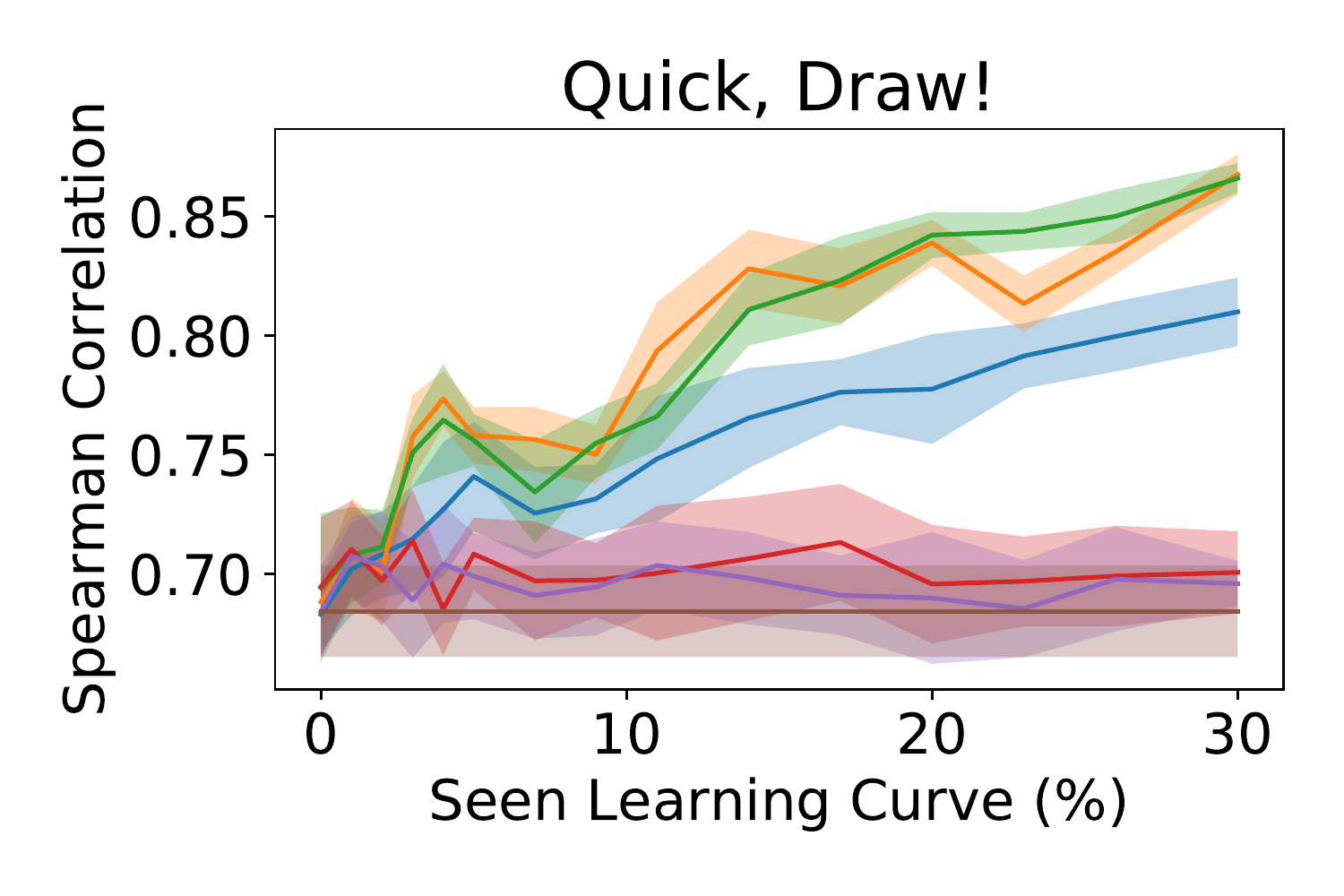}
\includegraphics[width=0.32\textwidth]{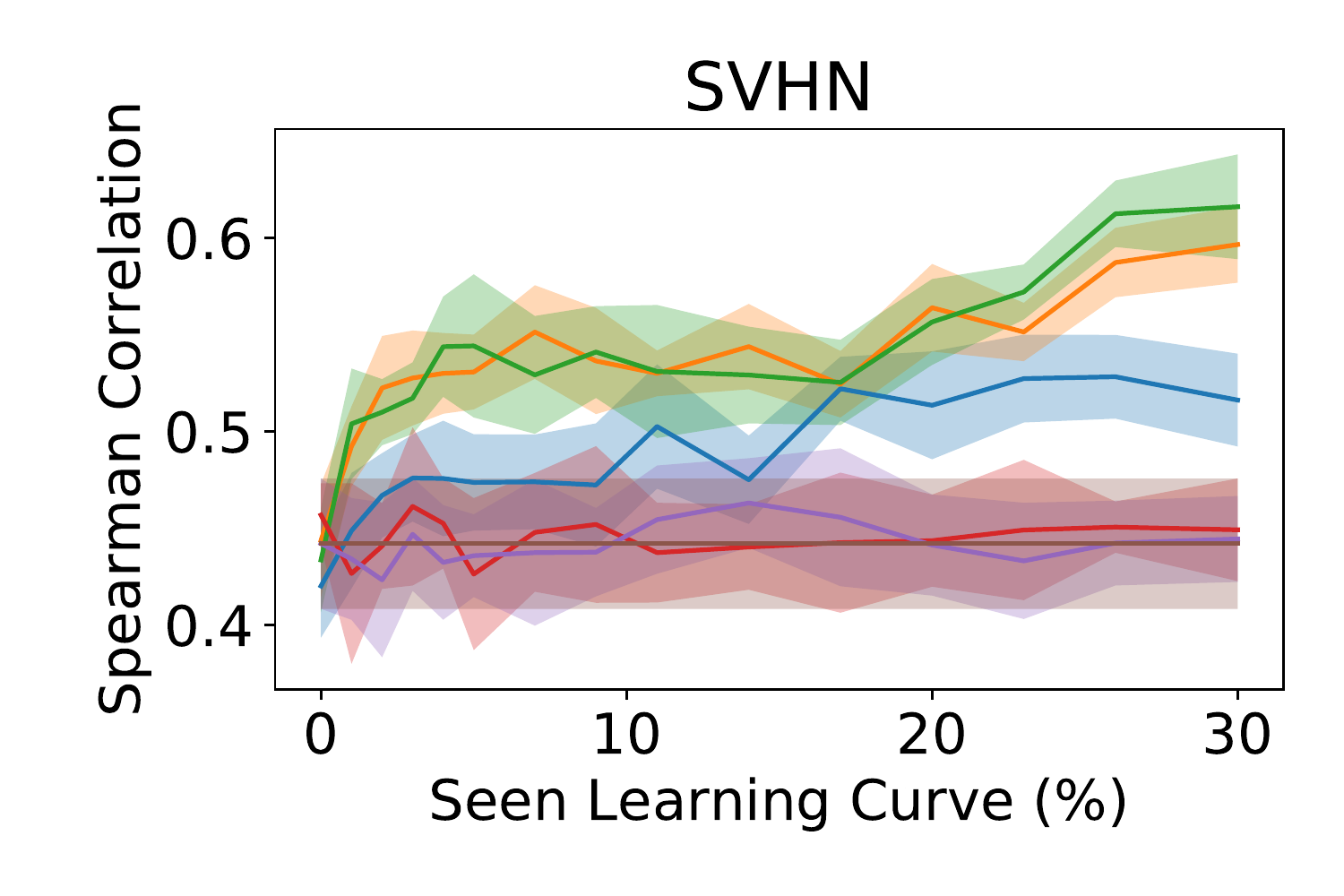}
\includegraphics[trim={14.7cm 1cm 0 1.5cm},clip,width=0.32\textwidth]{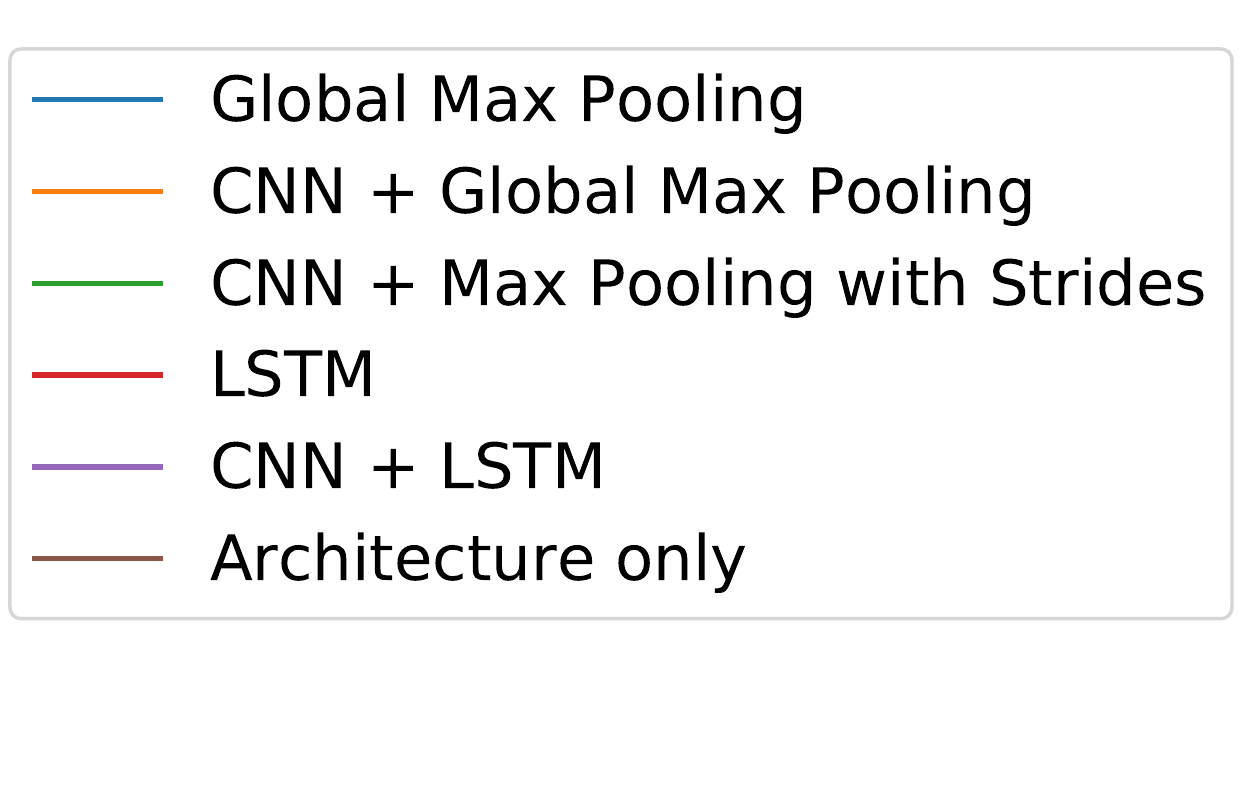}
\caption{Analysis of various different learning curve components in comparison to not using the learning curve at all.}
\label{fig:lcp-rank-results-lc-component}
\end{figure*}
\section{Ablation study: learning curve component}
One of the most important design decisions of \ourLCP{} was the learning curve component.
We experimented with several models that would generate the best learning curve embedding.
In the following discussion of different models, we assume that the learning curve represents the validation accuracy over time and thus higher values are better.

To begin with, passing the entire learning curve directly to be concatenated with architecture embedding would result in large number of predictors thereby overfitting.
It was also clear that the best or last value of the learning curve alone (assuming that the learning curve is constantly improving, i.e. it increases monotonically, it is the same) is a very good predictor.
After all, it is the only one used by methods like Hyperband and Successive Halving.
In our example, this would be achieved through a simple global max pooling layer.
Any further information regarding the development of the learning curve (improvement since epoch 1, 1st and 2nd order gradients, etc.) would, however, be disregarded.
Feature engineering would be one way to create such features but convolutions allow to automatically learn which of these features are helpful.
In Figure~\ref{fig:lcp-rank-results-lc-component} we take two different models with max pooling layer into account. The version with global max pooling layer reduces the number of predictors to one per filter, while the version with strides reduces the number to 4 per filter.
In our experiments, we did not notice a big difference between these two versions, but a significant improvement over the version that only uses the best value.

Furthermore, we tried in vain to apply LSTMs to this problem.
But both, LSTMs directly using the learning curve and using the output of the convolution did not achieve better results than if the learning curve had not been taken into account at all ("architecture only").

\section{Joint architecture and hyperparameter optimization}

As briefly discussed in the main paper, we want to show that \ourLCP{}
\begin{itemize}
    \item can be combined with optimization methods such as Tree-structured Parzen Estimator (TPE)~\citep{Bergstra2011}, Regularized Evolution (RE)~\citep{Real2019} and Reinforcement Learning (RL)~\citep{Zoph2017},
    \item can be applied to different search spaces (convolutional vs. fully connected neural networks) with and without hyperparameters, and
    \item work with different machine learning tasks (classification vs. regression).
\end{itemize}
For this reason we are carrying out an additional experiment on a publicly available tabular regression benchmark~\citep{Klein2019}.
This benchmark was created by performing a grid search with a total of 62,208 different architecture and hyperparameter settings for four different tabular regression datasets: slice localization, protein structure, parkinsons telemonitoring, and naval propulsion.
Each dataset is split into 60\% train, 20\% validation, and 20\% test.
Each architecture has two fully connected layers.
Settings vary with respect to the initial learning rate (0.0005, 0.001, 0.005, 0.01, 0.05, 0.1), the batch size (8, 16, 32, 64), the learning rate schedule (cosine or fixed), the activation per layer (ReLU or tanh), the number of units per layer (16, 32, 64, 128, 256, 512), and dropout per layer (0.0, 0.3, 0.6).

We are again following the leave-one-dataset-out cross-validation protocol:
We optimize the architecture and hyperparameter setting for one dataset and transfer the knowledge from the others.
100 settings per dataset are randomly selected as additional data for \ourLCP{}.

\subsection{Accelerating hyperparameter optimization}
\begin{figure*}
\centering
\includegraphics[width=0.4\textwidth]{early_stop_protein_structure_tpe.pdf}
\includegraphics[width=0.4\textwidth]{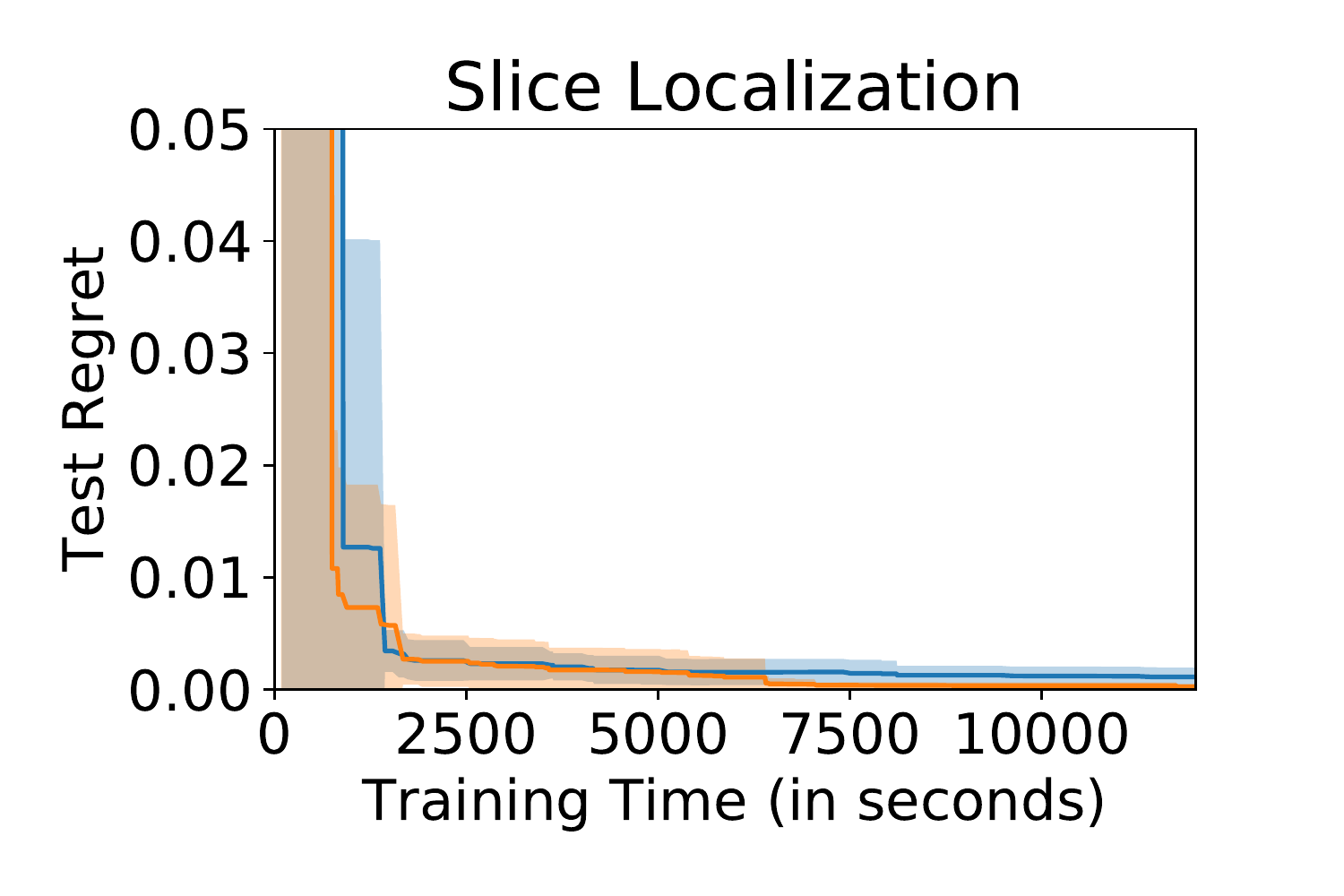}
\includegraphics[width=0.4\textwidth]{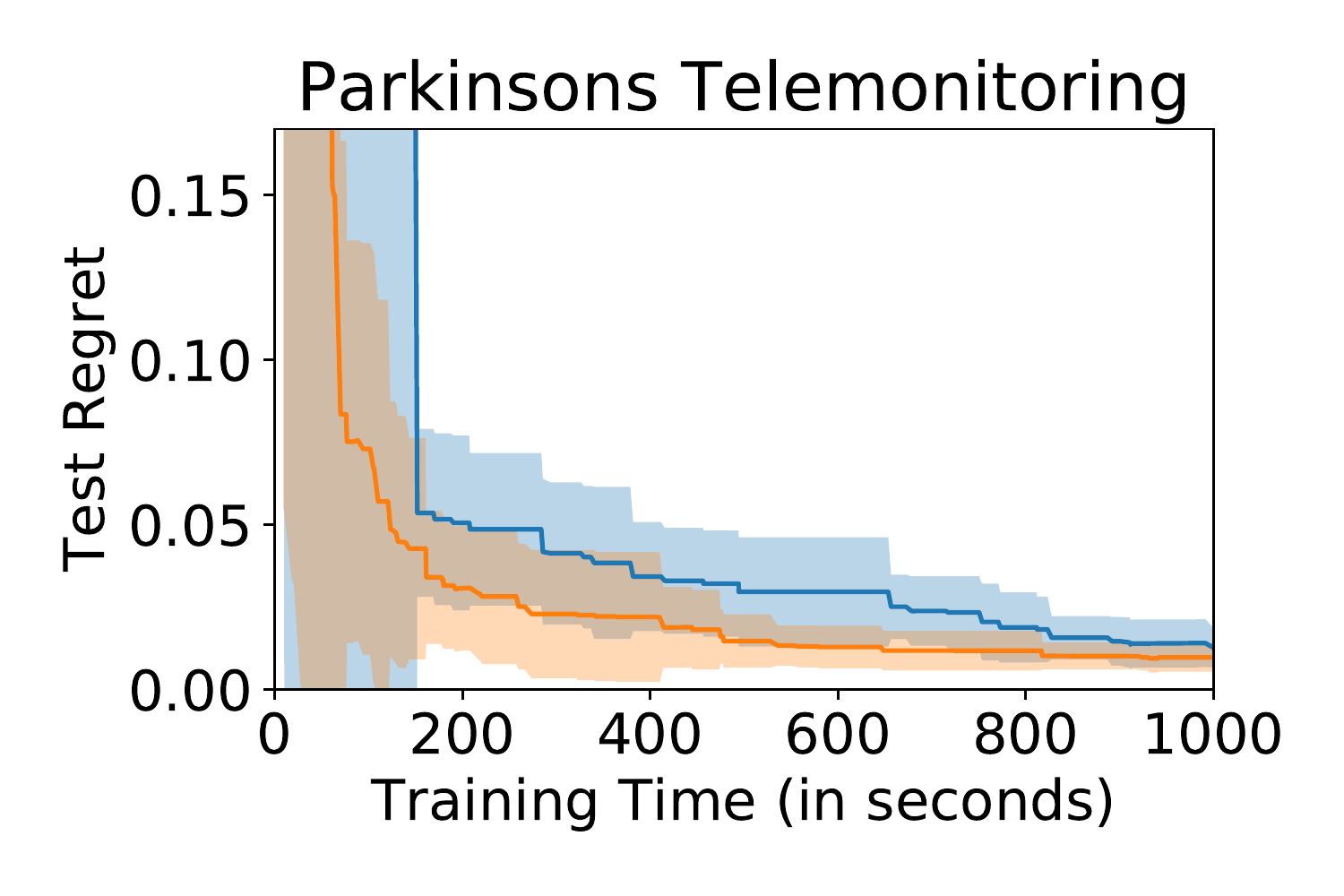}
\includegraphics[width=0.4\textwidth]{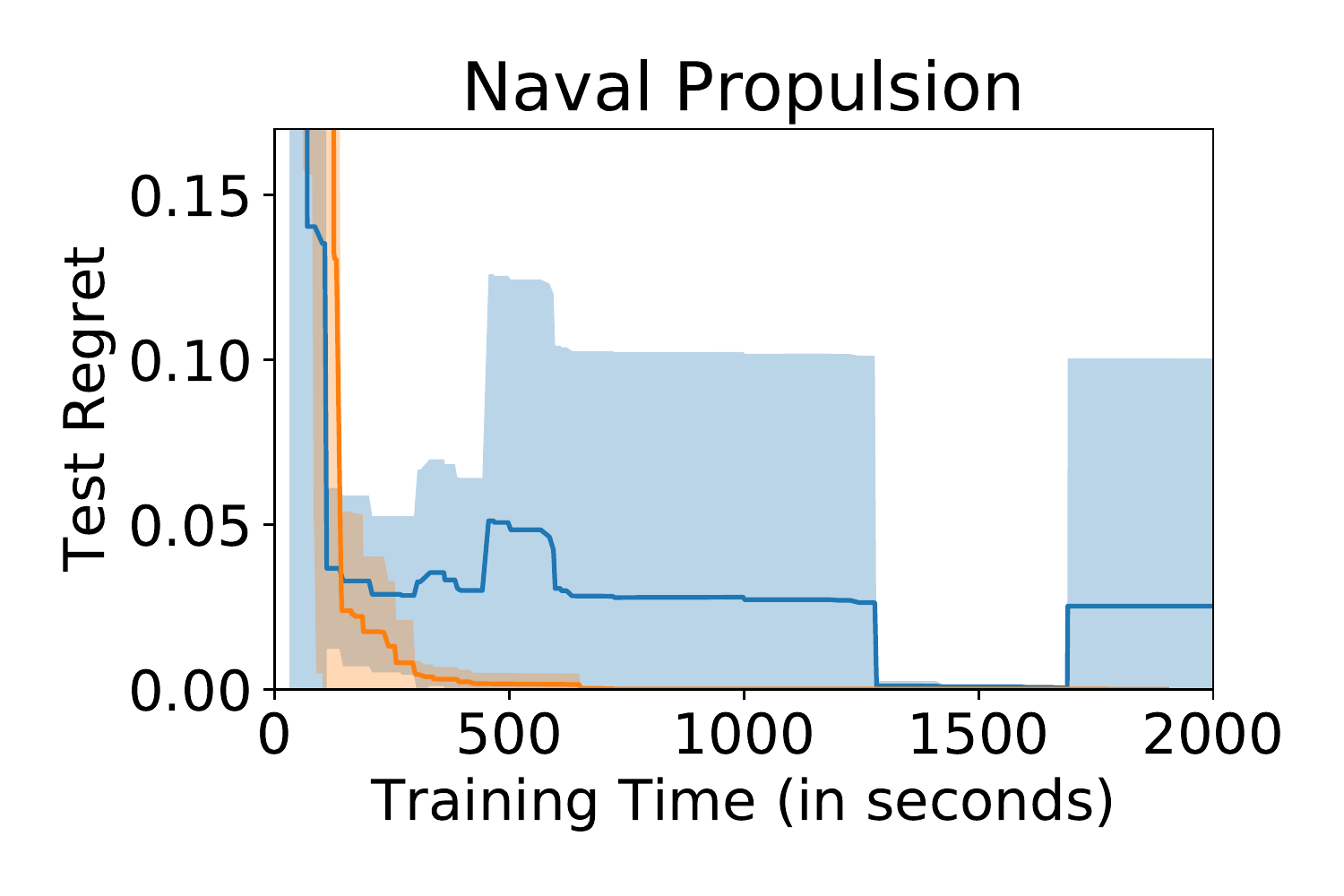}
\caption{TPE benefits from early termination on all datasets.}
\label{fig:early-stop-results-mlp-tpe}
\end{figure*}

\begin{figure*}
\centering
\includegraphics[width=0.4\textwidth]{early_stop_protein_structure_regularized_evolution.pdf}
\includegraphics[width=0.4\textwidth]{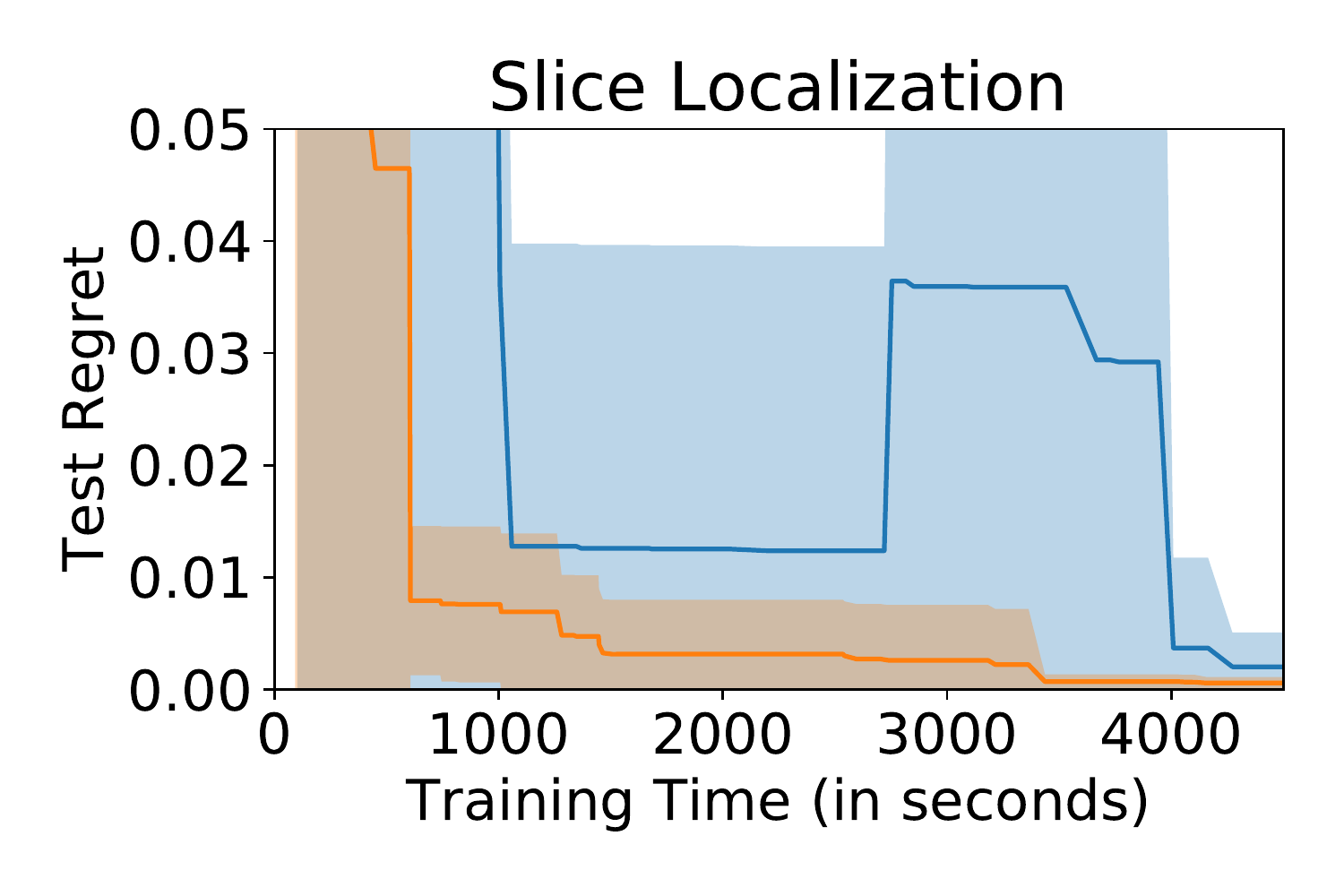}
\includegraphics[width=0.4\textwidth]{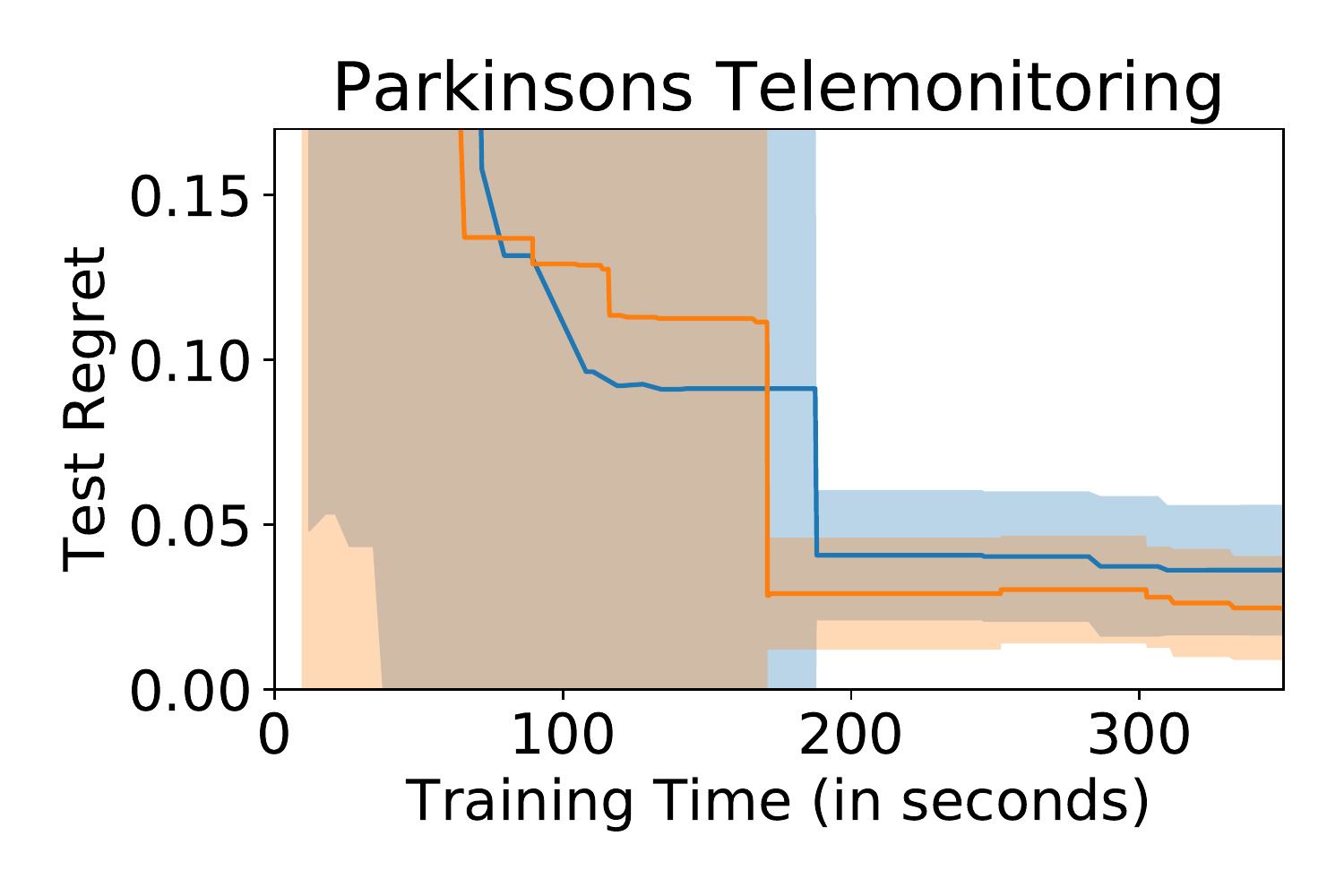}
\includegraphics[width=0.4\textwidth]{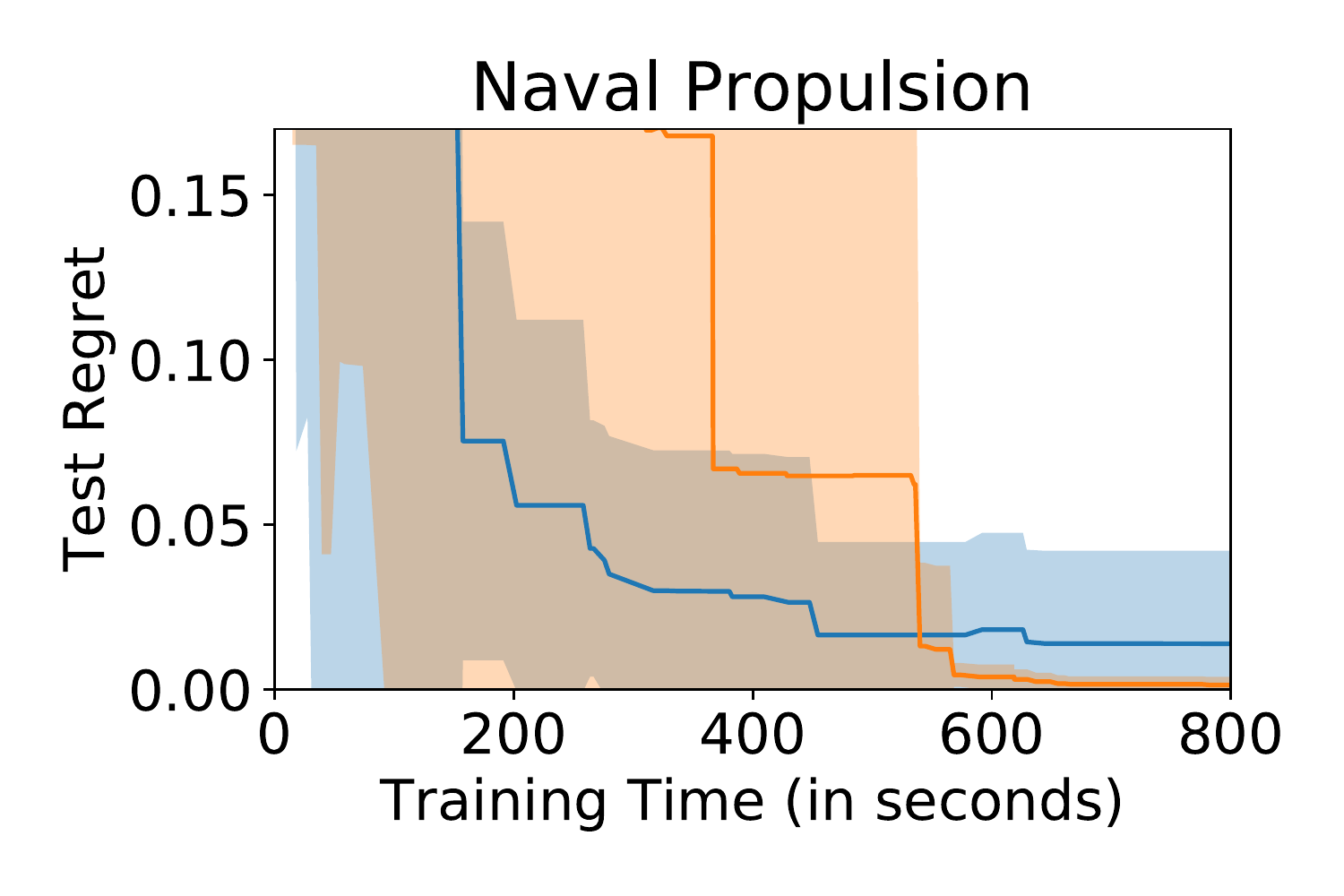}
\caption{Regularized Evolution benefits from early termination on all datasets.}
\label{fig:early-stop-results-mlp-regularized_evolution}
\end{figure*}

\begin{figure*}
\centering
\includegraphics[width=0.4\textwidth]{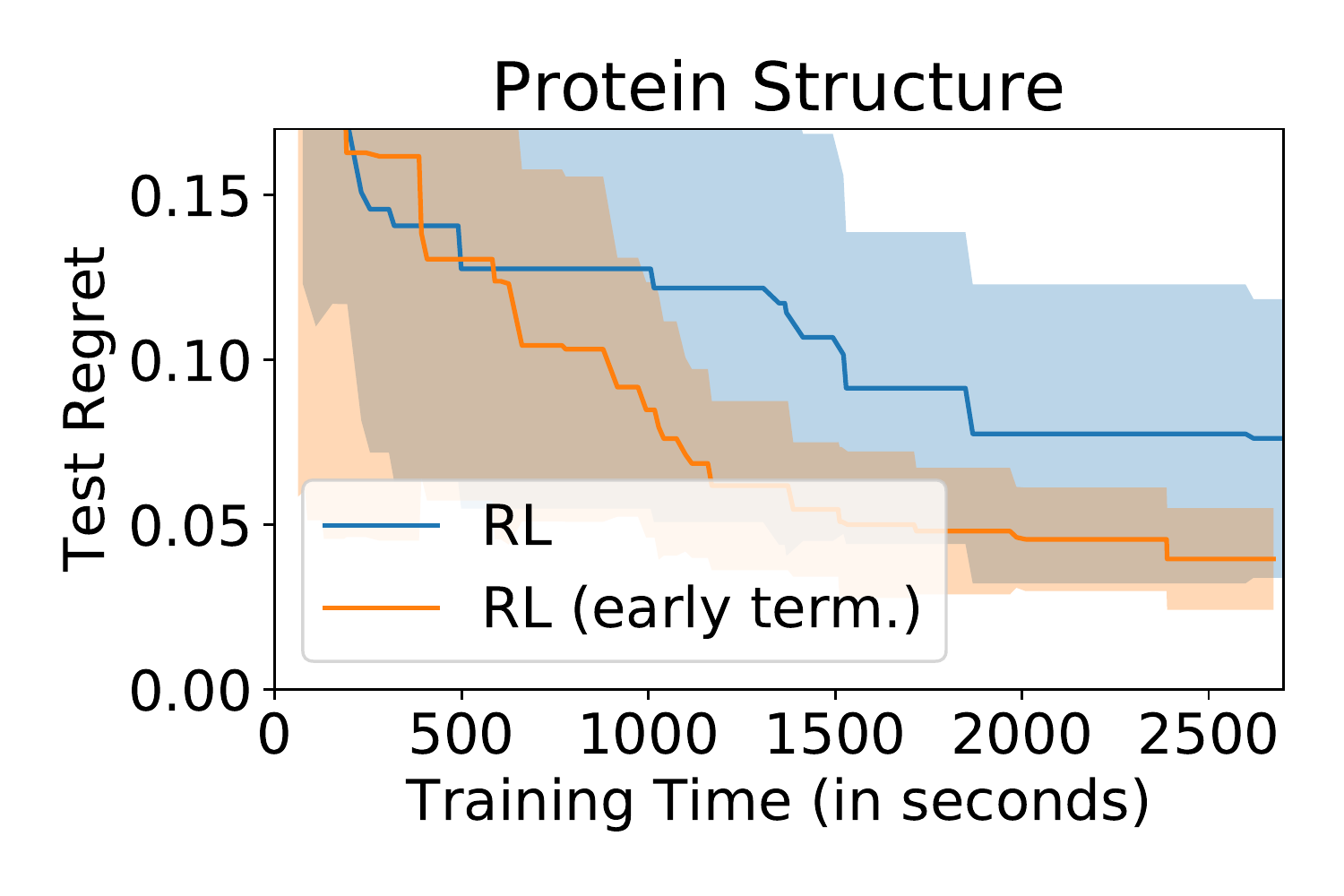}
\includegraphics[width=0.4\textwidth]{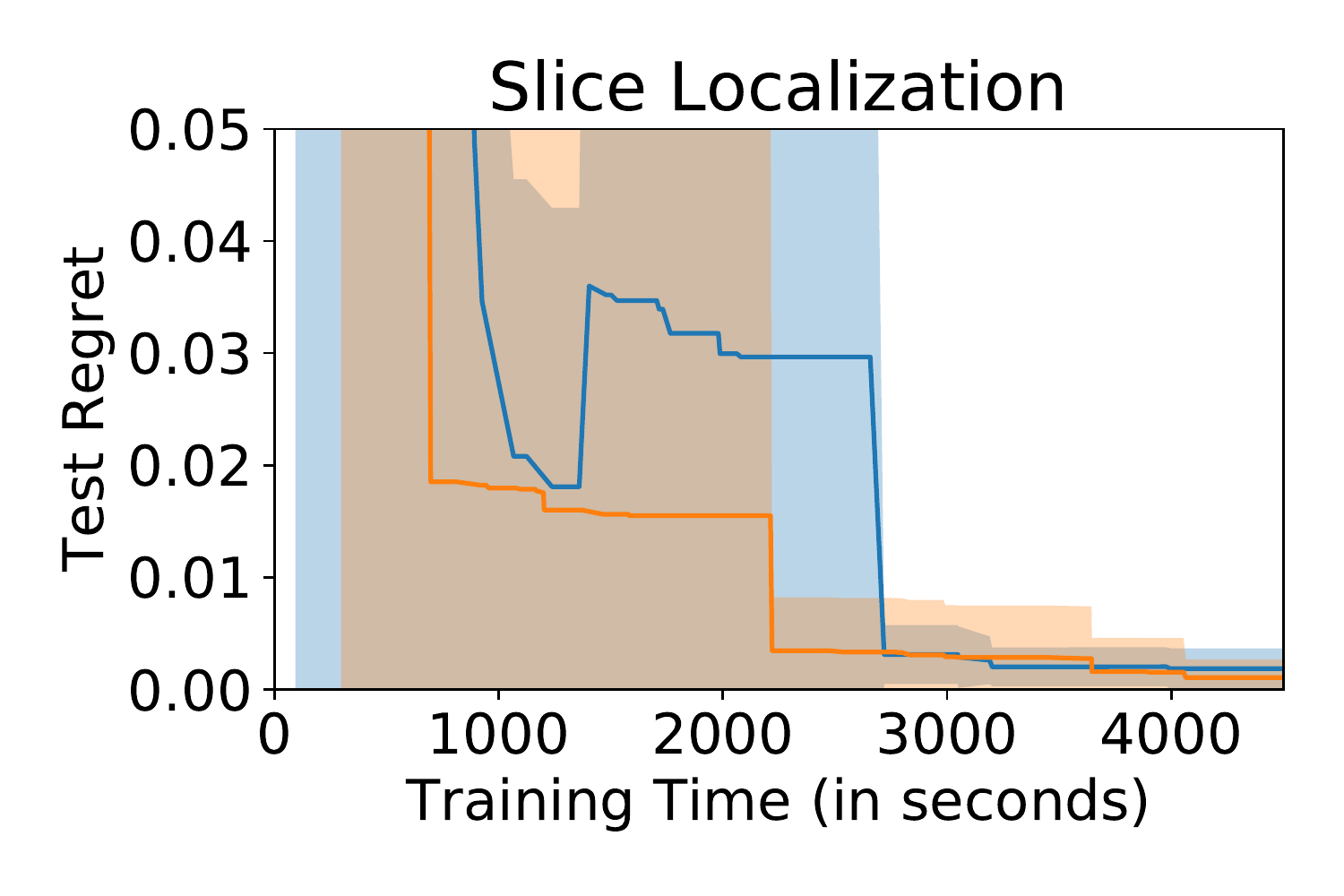}
\includegraphics[width=0.4\textwidth]{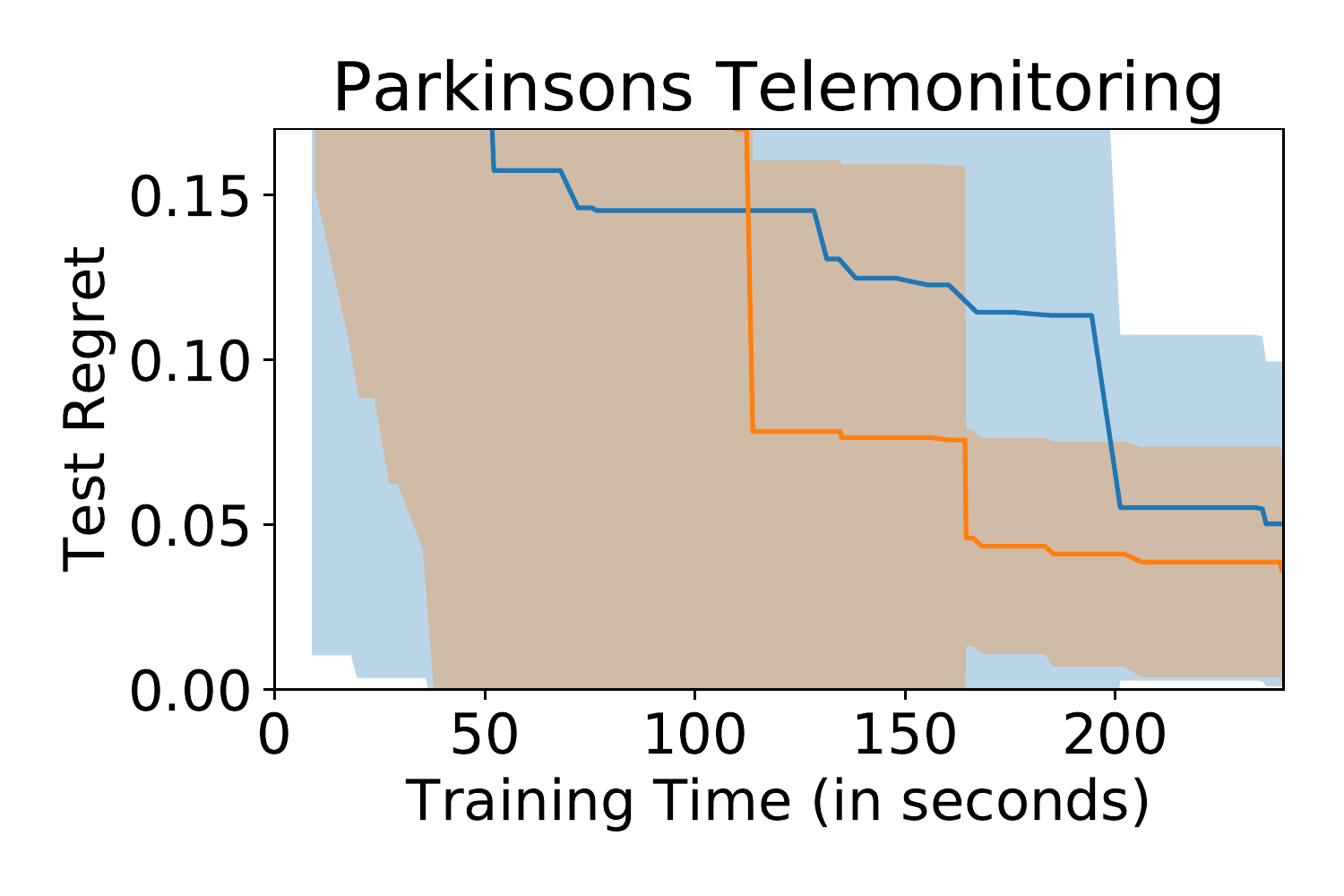}
\includegraphics[width=0.4\textwidth]{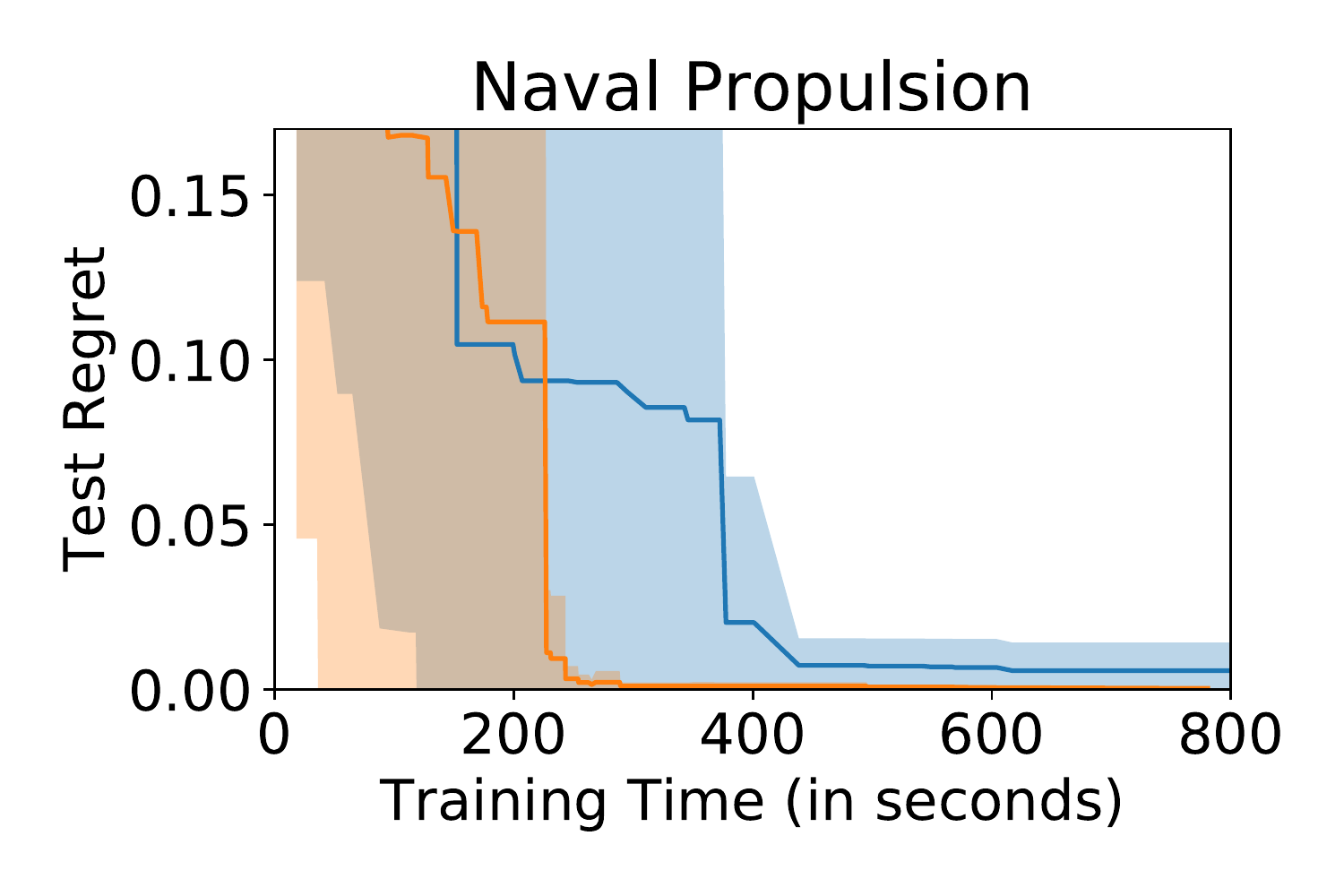}
\caption{Reinforcement Learning benefits from early termination on all datasets.}
\label{fig:early-stop-results-mlp-rl}
\end{figure*}

The goal is to minimize the mean squared error (MSE) by choosing the right architecture and hyperparameter settings.
In this experiment we compare the optimization methods TPE, RE and RL with a version with early termination using \ourLCP{}.
All experiments are repeated ten times.
Each optimization method can take up to 100 different settings into account.
We report results with respect to test regret following \citet{Klein2019}.
This number indicates the difference between test MSE actually achieved and best possible test MSE.
Finding the best setting will result in a regret of 0.
The setting chosen by an optimization method is based on the validation MSE.
It is therefore possible that the test regret may increase again.
This can be seen as an overfitting on the validation set.
In Figure~\ref{fig:early-stop-results-mlp-tpe} to \ref{fig:early-stop-results-mlp-rl} we report the results up to the search time required for the shortest of all repetitions.
We find that early termination generally improves the method.
Only in one case are they not better, but not worse either.

\subsection{Technical details}
In contrast to random search, Successive Halving or Hyperband, the intermediate performance is not sufficient for other optimization methods and the final performance of a model is required.
To take this into account, a simple change to \ourLCP{} is required.
An additional output layer is added that predicts the final performance in additional to the ranking score.
We continue to use the ranking score only to decide whether to terminate a run early or not.
The predicted final performance is only used in the event of early termination and is only used to provide feedback to the optimization method.
If a run is not terminated early, the actual final performance is used.
To ensure that the predicted final performance is in a reasonable range, we define a lower and upper bound.
We are now explaining them in the context of MSE so lower values are better and vice versa.
The lower bound is defined by the mean MSE observed for previous runs.
The motivation for this decision is that each terminated run should be below average.
We define the best observed MSE of the partial learning curve as an upper bound.
If the upper and lower bounds conflict, we prefer the upper bound because it is based directly on observed data.

\begin{figure}[t]
  \begin{center}
    \resizebox{0.2\columnwidth}{!}{\begin{tikzpicture}
\definecolor{colorSegment}{HTML}{ffec9e}
\definecolor{colorCell}{HTML}{ddff88}
\definecolor{colorReductionCell}{HTML}{ffdd88}
\definecolor{colorOp}{HTML}{a6f4fd}
\definecolor{colorBlock}{HTML}{bbee66}

\tikzstyle{boxstyle}=[rectangle,node distance=1.2cm,rounded corners=1ex,align=center]
\tikzstyle{opstyle}=[boxstyle,draw=black,fill=colorOp, minimum size=7mm,align=center]

  \node[opstyle](op1-2) {conv\\3x3};
  \node[opstyle](op1-1) [left of=op1-2] {max\\3x3};
  \node[opstyle](add1) [above=1 of $(op1-1)!0.5!(op1-2)$] {add};
  \draw[->]   (op1-1) -- (add1);
  \draw[->]   (op1-2) -- (add1);
  
\end{tikzpicture}}
  \end{center}
  \caption{One example block used in an architecture from the NASNet search space.}
  \label{fig:nasnet-block}
\end{figure}
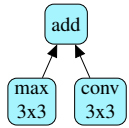
\section{Architecture representation}
In this work we consider two different search spaces.
The NASNet search space that takes CNNs~\citep{Zoph2018} into account and a search space for FCNNs~\citep{Klein2019}.
In the NASNet search space, an architecture is completely described by two cells, each consisting of several blocks.
The entire architecture is defined by selecting the design of all blocks.
A block is designed by selecting two operations and their corresponding inputs and how these two operations are combined.
A sample block is shown in Figure~\ref{fig:nasnet-block}.
We follow \citet{Luo2018} and model every combination of input and operation through three embeddings.
The first embedding specifies the input choice, the second the type of operation (convolution, maximum pooling, etc.), and the third the kernel size.
In this way, an architecture with two cells, each with five blocks, is clearly described by a sequence of 60 decisions.

We describe the FCNNs in a very similar way.
For each layer we learn an embedding for the activation function, the dropout rate and the number of units.
The batch size, the learning rate (after log transformation) and the learning rate schedule (after one-hot encoding) are taken into account as other hyperparameters.

\end{document}